\documentclass[preprint,12pt]{elsarticle}

\usepackage{amssymb}
\usepackage{amsmath}
\usepackage{multirow}
\usepackage{verbatim}
\usepackage{booktabs}
\usepackage{dsfont}
\usepackage{graphicx}
\usepackage{colortbl}
\usepackage{algorithmic}
\usepackage{algorithm}

\journal{Nuclear Physics B}

\begin{document}

\begin{frontmatter}



\title{Knowing Beyond the Known: Reinforced Knowledge Specification for Multi-Label Class-Incremental Learning}


\author[1,3]{Aoting Zhang} 
\author[2]{Dongbao Yang\corref{cor1}} 
\author[5]{Chang Liu} 
\author[4]{Xiaopeng Hong} 
\author[1,3]{Can Ma}
\author[2]{Yu Zhou\corref{cor1}} 
\cortext[cor1]{Corresponding authors.}

\affiliation[1]{organization={Institute of Information Engineering, Chinese Academy of Sciences},
            city={Beijing},
            postcode={100085}, 
            country={China}}
\affiliation[2]{organization={Nankai University},
            city={Tianjin},
            postcode={300350}, 
            country={China}}
\affiliation[3]{organization={School of Cyber Security, University of Chinese Academy of Sciences},
            city={Beijing},
            postcode={101408}, 
            country={China}}
\affiliation[4]{organization={Harbin Institute of Technology},
            city={Harbin},
            postcode={150001}, 
            country={China}}
\affiliation[5]{organization={Tsinghua University},
            city={Beijing},
            postcode={100084}, 
            country={China}}

\begin{abstract}
{Existing class-incremental learning methods struggle in multi-label scenarios (MLCIL) due to the inherent contradiction of learning objectives arising from co-occurring and incomplete labels. We argue that the core obstacle is the model's ambiguous boundary between known and unknown knowledge, which undermines historical knowledge retention, complicates current task learning, and limits adaptability to future concepts. To address this, we propose KBK (Knowing Beyond the Known), a reinforced knowledge specification framework that explicitly models what is known or not to unify historical, current, and prospective learning. Specifically, to clarify known knowledge, we develop a hierarchical feature purification module that disentangles fine-grained class-specific features from global features, where high-level semantic abstraction is reinforced with low-level visual features. Additionally, an uncertainty-aware recall enhancement strategy suppresses unreliable predictions based on distribution priors, improving the quality of historical recall. For probing the unknown, KBK leverages semantic correlations to synthesize informative unknown features under co-occurring, preserving embedding space for future learning. Furthermore, to mitigate heterogeneous forgetting, we design a category-balanced gradient compensation loss that dynamically reweights gradient backpropagation according to forgetting speeds. Experiments on multiple benchmarks validate the effectiveness and robustness of KBK, which surpasses prior best methods by 2.7\% in Avg. Acc on MS-COCO B0-C10 setting even without any replay buffers.}
\end{abstract}


\end{frontmatter}


\section{Introduction}

Class-incremental learning (CIL) is designed to incrementally incorporate novel classes into a unified model without compromising its ability to recognize previously learned ones. Many efforts have focused on mitigating catastrophic forgetting, a challenge exacerbated by the unavailability of old data during continual updates. Most existing studies, however, are designed for single-label class-incremental learning (SLCIL), which assumes each image is associated with only one label. In reality, real-world images often contain multiple semantically meaningful objects, such as cars, pedestrians, and buses in urban scenes. This has spurred growing interest in multi-label class-incremental learning (MLCIL)~\cite{dong2023knowledge}, which focuses on correctly assigning images to multiple classes that appear across successive sessions.
\begin{figure}[t]
  \centering
  \includegraphics[width=0.86\linewidth]{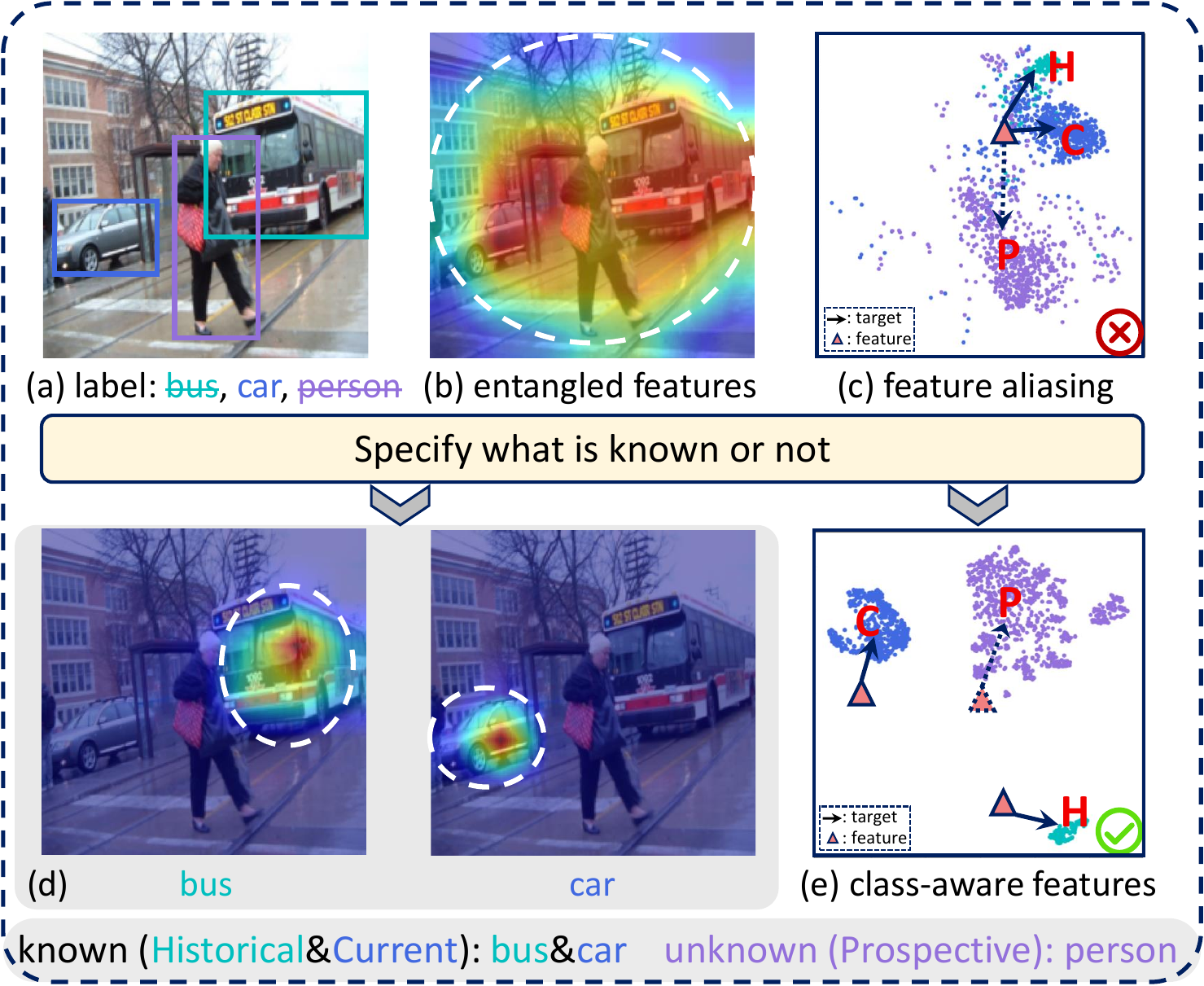}
  \caption{The contradiction of learning objectives arises from the model's inability to distinguish known and unknown knowledge. Current model fails to effectively recall prior known knowledge due to (a) the absence of historical labels, while (b) unknown classes' attention is inadvertently overlapped with known classes, and known classes are also entangled, resulting in (c) feature aliasing and contradictory learning objectives. By specifying what is known or not, (d) fine-grained class-aware features are focused, leading to (e) enhanced inter-class discriminability, alleviating the contradiction.}
  \label{fig:krt_ckat_attn}
\end{figure}

MLCIL presents a unique challenge named partial labeling--{individual images may contain objects from previously learned (historical), currently introduced, and even yet-to-be-seen (future) classes, while supervision is provided only for current session (e.g., only ``car'' is labeled in Fig.~\ref{fig:krt_ckat_attn} (a)), and other co-occurring classes such as ``bus'' and ``person'' remain unannotated.}
Naively applying anti-forgetting remedies from single-label CIL, such as knowledge distillation and exemplar replay, proves ineffective in MLCIL, since these methods fail to reconcile the conflicting learning objectives inherent to incomplete labels. Specifically, MLCIL introduces a tripartite tension among three critical learning goals--preserving previously acquired knowledge, optimizing classification for current classes, and preparing for subsequent tasks. To preserve old knowledge, recent approaches employ pseudo-labeling of prior classes and multi-knowledge retention strategies. For example, KRT~\cite{dong2023knowledge} proposes a knowledge restoration and transfer framework to compensate for missing labels of known classes. APPLE~\cite{song2024overcoming} adopts adaptive pseudo-label strategy to increase
the label information in the training data.

{Despite these advancements, existing approaches mainly focus on historical knowledge retention or current-label completion, while the ambiguous boundary between reliable known knowledge and uncertain unknown content remains underexplored. Under partial labeling and label co-occurrence, unlabeled future classes may appear in current-session images and be mistakenly absorbed into known-class regions. This leads to representation entanglement, where the model activates unseen-class features as if they belong to known classes.}
As illustrated in Fig.~\ref{fig:krt_ckat_attn} (b), although not ``knowing'' the person class, the model erroneously focuses on the corresponding region and conflates person features with those of known classes such as car and bus. In addition, insufficient ``knowing'' ability--manifested as weak discrimination between past and current classes--exacerbates feature entanglement under label co-occurrence, blurring task boundaries and intensifying learning conflicts. As a result, not only is future learning compromised, but the recall of previous knowledge is also hindered due to the absence of explicit supervision, thus exacerbating forgetting.

To address this, we propose Knowing Beyond the Known (KBK), a reinforced knowledge specification framework designed to explicitly model what is known or not to accommodate historical, current, and prospective knowledge. \textit{For clarifying known knowledge}, we adopt a hierarchical feature purification module to capture fine-grained class-aware representations while suppressing irrelevant noise. Specifically, high-level semantic features are reinforced with intermediate visual cues via residual fusion, enriching spatial details while preserving semantic abstraction. Each class is presented by a learnable embedding, which interacts with fused features via an attention mechanism to locate relevant regions. It not only enhances the discrimination of co-occurring objects but also allows the embedding space to be seamlessly extended to new classes. To improve the reliability of knowledge recall, we incorporate an uncertainty-aware recall enhancement strategy that adaptively filters out low-confidence and high-uncertainty predictions based on historical distribution priors, enhancing the quality of recalled supervision.
\textit{For probing unknown knowledge}, KBK leverages semantically correlated absent-class embeddings to synthesize prospective features that emulate potential future classes, enabling the model to reserve embedding space for unseen concepts while strengthening discrimination among current classes.
Furthermore, we design a \textit{category-balanced gradient compensation loss} to mitigate the heterogeneous forgetting of old classes by reweighting gradient backpropagation. The gradients of heavily forgotten classes are amplified and those of well-remembered ones are suppressed, effectively rectifying imbalanced gradient distributions.

To summarize, our major contributions are as follows:
\begin{itemize}
    \item {We propose KBK, a unified known/unknown knowledge-specification framework for MLCIL. Instead of treating historical forgetting, current-class learning, and future-class interference as isolated problems, KBK explicitly coordinates \emph{historical}, \emph{current}, and \emph{prospective} knowledge by specifying reliable known knowledge and uncertain unknown content across incremental sessions.}
    \item {Under this principle, we design a Hierarchical Feature Purification module to extract class-aware known representations from co-occurring objects, and an Uncertainty-aware Recall Enhancement strategy to recover reliable historical supervision from partially labeled current-session data.}
    \item {To model prospective knowledge, we introduce Semantic-guided Probing Unknown, which synthesizes unknown features from absent-class responses and assigns unexplained content to an explicit unknown direction. We further develop Category-balanced Gradient Compensation to alleviate heterogeneous forgetting by balancing class-wise optimization signals.}
    \item Evaluation on diverse experimental protocols confirms that KBK attains superior performance while diminishing catastrophic forgetting.
\end{itemize}

{This manuscript substantially extends the conference
work~\cite{zhang2025specifying} 
by introducing several major technical contributions and a much broader empirical evaluation:
1) The feature purification module is advanced from a flat structure to a hierarchical paradigm that facilitates bidirectional fusion between high-level semantic abstractions and intermediate spatial cues, thereby yielding more discriminative and fine-grained class-aware representations.
2) To enhance the reliability of historical recall under partial labeling, we incorporate uncertainty-aware pseudo-labeling that adaptively filters out high-uncertainty predictions.
3) Moving beyond random feature interpolation, we introduce a context-aware synthesis mechanism that exploits latent semantic relations to synthesize informative unknown features, reserving embedding space for future concepts.
4) We address the long-standing issue of heterogeneous forgetting by designing a category-balanced gradient compensation loss that reweights the gradient propagation based on each class's forgetting tendency.
5) We provide an exhaustive evaluation across diverse and more rigorous experimental protocols, including long-term incremental sequences and non-overlapping session designs. Notably, the enhanced KBK achieves a substantial performance leap, specifically a 2.7\% gain in Avg. Acc over the conference version (HCP) on the challenging Split-VOC B10-C5 setting. Expanded ablation studies and interpretability analyses further confirm the model's enhanced ability to maintain distinct representations for successive incremental tasks.}

\section{Related Work}

\subsection{Single-Label Incremental Learning}

SLCIL focuses on incrementally integrating novel classes from a data stream while retaining competence on previously learned ones. In scenarios without access to former data, existing solutions can be categorized into three main types to address forgetting. 
\textit{Regularization-based methods} constrain the update of important parameters or responses from previous tasks. EWC~\cite{kirkpatrick2017overcoming} estimates parameter importance with Fisher information, while later works refine importance estimation and optimization~\cite{schwarz2018progress, aljundi2018memory}. Knowledge-distillation methods, such as LwF~\cite{li2017learning} and PODNet~\cite{douillard2020podnet}, preserve previous model behaviors through logit- or feature-level distillation.
\textit{Rehearsal-based methods} store a small number of exemplars from old classes and replay them during later training. Representative methods include ER~\cite{riemer2018learning}, iCaRL~\cite{rebuffi2017icarl}, DER++~\cite{buzzega2020dark}, and BiC~\cite{wu2019large}, which improve old-class retention through exemplar replay, distillation, or bias correction.
\textit{Architectural-based methods} expand or adapt model structures for new tasks~\cite{sun2023exemplar, liu2025class}. DER~\cite{yan2021dynamically} introduces task-specific feature extractors, DyTox~\cite{douillard2022dytox} uses task tokens, and recent prompt-based methods~\cite{wang2022learning, wang2022dualprompt} learn task-adaptive prompts on pre-trained ViTs. Unlike these SLCIL methods, our work focuses on MLCIL, where multiple co-occurring labels and partial annotations require explicit specification of known and unknown knowledge.

\subsection{Multi-Label Classification}
Multi-Label Classification (MLC) remains challenging compared with single-label classification. A straightforward solution treats each category independently and trains multiple \textit{one-vs-rest} binary classifiers. However, this independence assumption ignores inter-label correlations and spatial co-occurrence among objects, both of which are crucial in MLC. To leverage label structure, prior works adopt sequence models, typically RNNs, to model inter-label relationships, yet they are sensitive to label ordering and often difficult to optimize. Graph-based methods instead construct a label graph and apply Graph Convolutional Networks (GCNs) to capture relationships, but when co-occurrence statistics are sparse, they risk encoding spurious correlations. Beyond architecture, loss design also matters--the standard binary cross-entropy encounters severe imbalance between positive and negative labels in MLC. Asymmetric Loss (ASL) addresses this by dynamically down-weighting easy negatives and applying hard thresholding. {Recent studies further improve MLC by revisiting label correlation modeling and class-aware representation learning. For example, transformer-based methods leverage long-range visual-label dependencies to boost multi-label recognition~\cite{li2023patchct}. More recent works model label correlations with latent context~\cite{chen2024modeling}, enhance label-specific visual representations with semantic guidance~\cite{zhu2024semantic}, and pursue causal label correlations to suppress spurious contextual dependencies~\cite{chen2024pursuit}. These methods improve static MLC from the perspectives of context modeling, semantic representation learning, and robust label-correlation modeling.}

{However, most existing MLC methods assume a fixed label space and jointly observed training data. They are not directly designed for the class-evolving and partially labeled setting of MLCIL, where historical and prospective classes may appear in current-session images without annotations. Directly extending static MLC methods to this setting can suffer from catastrophic forgetting, feature aliasing, and interference from unlabeled co-occurring objects. In contrast, KBK explicitly specifies known and unknown knowledge to coordinate historical retention, current-class discrimination, and future-class preparedness.}

\subsection{Multi-Label Class-Incremental Learning}
Multi-Label Class-Incremental Learning (MLCIL) extends continual recognition to settings where images contain multiple co-occurring objects and only a subset of labels is annotated at each session. Recent efforts tackle this problem from complementary angles. On the memory side, PRS~\cite{kim2020imbalanced} devises a replay sampling strategy to alleviate the buffer class imbalance, and OCDM~\cite{liang2022optimizing} adopts a greedy update policy for fast and effective exemplar maintenance. To model inter-label structure, AGCN~\cite{du2023multi} employs GCNs to maintain consistent label relations across sessions. KRT~\cite{dong2023knowledge} tackles label absence with a knowledge restoration and transfer framework. {CSC~\cite{du2024confidence} constructs cross-task label relationships and calibrates confidence to curb over-confident predictions. 
Although these methods improve MLCIL from the perspectives of replay, label-relation modeling, pseudo-labeling, confidence calibration, and rebalancing, they focus on preserving historical knowledge or improving current-session learning. The ambiguity between reliable known knowledge and uncertain unknown content under label co-occurrence is still not explicitly modeled. In contrast, KBK formulates MLCIL from a known/unknown knowledge-specification perspective and jointly coordinates \emph{historical}, \emph{current}, and \emph{prospective} knowledge, thereby alleviating the conflict among knowledge retention, current-class discrimination, and future-class preparedness.}

\subsection{Incremental Learning in Related Vision Tasks}
Incremental object detection (IOD) represents a specialized form of MLCIL extended to detection, where models continually localize and recognize novel classes while retaining prior ones~\cite{yang2022rd, dong2023class, zhang2025dca}. 
Shmelkov et al.~\cite{shmelkov2017incremental} introduce IOD for the two-stage object detector and distill the output of Fast R-CNN. 
MVCD~\cite{yang2022multi} preserves multi-dimensional correlations within feature maps to stabilize updates. Replay-based strategies are also explored, either through feature replay or exemplar storage~\cite{yang2023one}.

{Incremental semantic segmentation (ISS) is required to maintain pixel-wise predictions for previously seen classes. SDR~\cite{Michieli_2021_CVPR} leverages prototype matching and contrastive learning to enforce consistency in the latent space. PLOP~\cite{Douillard_2021_CVPR} employs multi-scale feature distillation. ALIFE~\cite{oh2022alife} introduces feature replay combined with an adaptive regularizer to balance accuracy and efficiency.}

Despite methodological parallels, methods in IOD and ISS cannot be directly applied to MLCIL, as they rely on task-specific architectures and supervision (bounding boxes or pixel-level annotations). In contrast, our work targets settings with \emph{only image-level labels}, underscoring the unique challenges of MLCIL and the necessity of tailored solutions.

\section{Proposed Method}
\subsection{Problem Formulation}
MLCIL aims to develop a unified model that can correctly identify all classes encountered throughout the learning stream. We consider $T$ incremental sessions. Let $\mathcal{D}=\{(x,y)\}$ denote the full dataset, where $x$ is an image and $y\in\{0,1\}^{|\mathcal{C}|}$ is a multi-hot label vector over the global class set $\mathcal{C}$. we partition $\mathcal{C}$ into $T$ disjoint blocks $\{\mathcal{C}^1,\dots,\mathcal{C}^T\}$ in lexicographic order and induce a corresponding partition of the data $\{\mathcal{D}^1,\dots,\mathcal{D}^T\}$, with $\mathcal{C}^m\cap \mathcal{C}^n=\emptyset$ for $m\neq n$ and $\bigcup_{t=1}^T \mathcal{C}^t=\mathcal{C}$. At session $t$, the learner has access only to $\mathcal{D}^t$ with the {current} label space $\mathcal{C}^t$. Different from SLCIL, images in $\mathcal{D}^t$ may simultaneously contain objects from past classes $\mathcal{C}^{1:t-1}$ and future classes $\mathcal{C}^{t+1:T}$. {Accordingly, we refer to $\mathcal{C}^{1:t-1}$, $\mathcal{C}^t$, and $\mathcal{C}^{t+1:T}$ as \emph{historical knowledge}, \emph{current knowledge}, and \emph{prospective knowledge}, respectively. Since only labels in $\mathcal{C}^t$ are observed at session $t$, the session-wise known knowledge is defined as the union of historical and current knowledge, i.e.,
$\mathcal{K}_{\mathrm{known}}^t = \mathcal{C}^{1:t-1}\cup \mathcal{C}^t,$
whereas prospective knowledge remains outside the observed label space and thus appears as currently unknown content.}
However, annotations outside $\mathcal{C}^t$ are {unavailable}, that is, labels in $\mathcal{C}\setminus \mathcal{C}^t$ are treated as absent. After completing session $t$, the model is evaluated on all classes $\mathcal{C}^{1:t}=\bigcup_{i=1}^t \mathcal{C}^i$, producing prediction $\hat{y}^{t}(x)\in[0,1]^{|\mathcal{C}^{1:t}|}$ for each image. The objective is to continually expand the label space with minimal degradation of previous classes.
\begin{figure}[t]
  \centering
  \includegraphics[width=\linewidth]{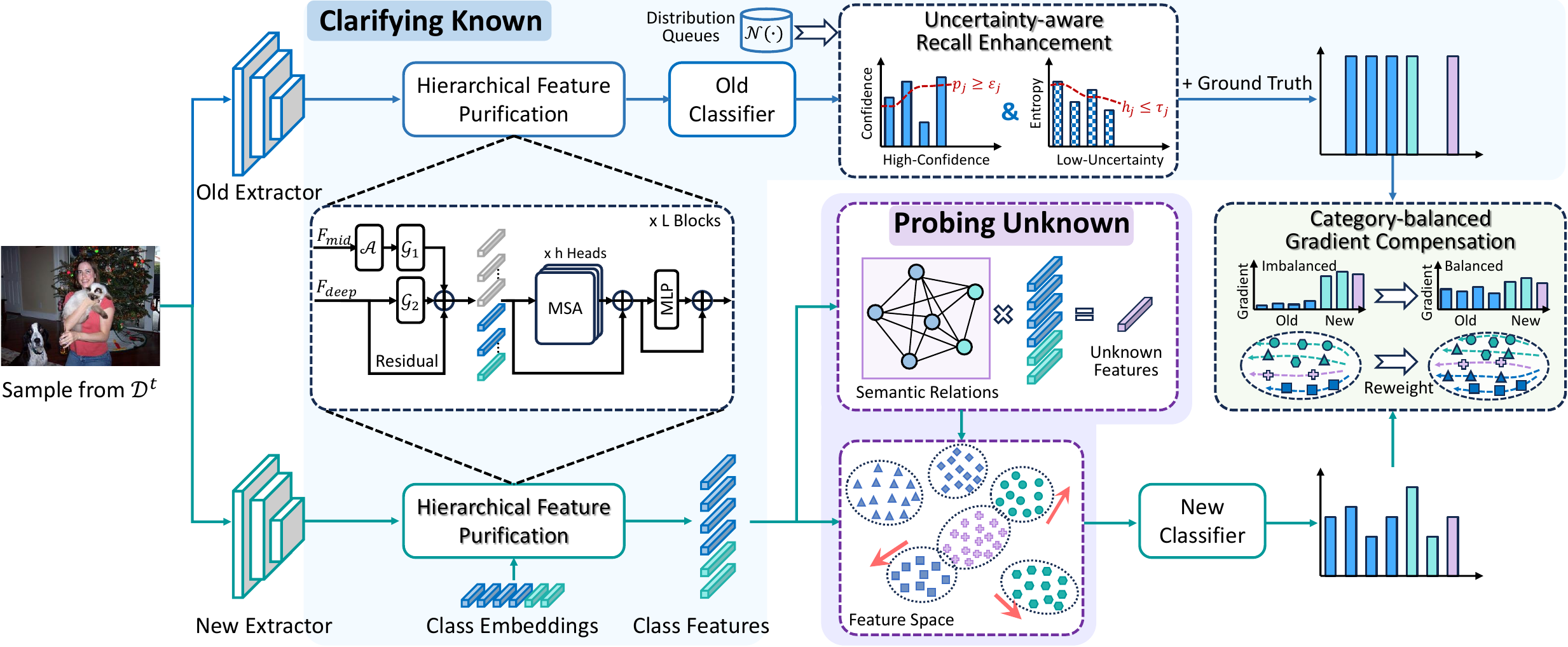}
  \caption{Framework of KBK. {KBK coordinates historical, current, and prospective knowledge under a unified known/unknown knowledge-specification principle. HFP and URE clarify reliable historical and current knowledge by extracting class-aware features and recalling high-confidence, low-uncertainty historical supervision. SPU probes unknown knowledge by synthesizing semantic-guided unknown features, while CGC balances historical retention and current learning through class-wise gradient compensation.}}
  \label{framework}
\end{figure}

\subsection{Overview}
{Figure~\ref{framework} depicts the proposed KBK framework, which is organized around a unified known/unknown knowledge-specification principle. The goal is to identify reliable known knowledge from historical and current classes, while preventing uncertain unknown content from being absorbed into known-class representations. In this way, KBK jointly coordinates historical retention, current-class discrimination, and prospective preparedness under the MLCIL setting.}
Concretely, {to make the \emph{known} more discriminative}, KBK introduces a hierarchical feature purification module, in which for each class, a dedicated embedding is trained to extract fine-grained class features from a fused representation that combines reinforced high-level semantic abstractions with complementary low-level visual signals. This hierarchical fusion reduces feature aliasing across sessions and supports flexible extension to new classes by continually adding class embeddings. In addition, we employ an uncertainty-aware recall enhancement strategy that combines confidence with entropy-guided uncertainty to filter unreliable pseudo labels, thus {improving the reliability of historical recall and mitigating uneven forgetting across classes}. {To probe the unknown knowledge}, KBK leverages semantic relations to weight absent-class embeddings and synthesize informative unknown features, which push non-target representations away, leading to more compact class boundaries for known classes while reserving space for future learning. Finally, to balance heterogeneous forgetting of old classes, we introduce a category-balanced gradient compensation loss that dynamically reweights gradient propagation based on forgetting dynamics.

\subsection{Clarifying Known Knowledge}
Here we introduce the hierarchical feature purification module and outline how it adapts to multi-label incremental scenarios. Subsequently, we study heterogeneous class forgetting and develop an uncertainty-aware recall enhancement that incorporates historical distributional priors. 

{\textit{1) Hierarchical Feature Purification.}}
To eliminate cross-session feature aliasing caused by noise and impurities from non-target features, we design a hierarchical feature purification module, which fuses shallow spatial and deep semantic information via residual integration, then uses class embeddings as queries to disentangle fine-grained class-specific signals from multi-class entanglement. Compared with other methods~\cite{dong2023knowledge, du2024confidence}, our approach ensures distinctive representations for each class by incorporating complementary information from multiple feature levels, enabling parallel prediction of both historical and current classes.

For a given sample from dataset $\mathcal{D}^{t}$, we extract features from the intermediate and last layers of the backbone network, denoted as $F_{mid}\in\mathbb{R}^{H_1\times W_1\times d_1}$ and $F_{deep}\in\mathbb{R}^{H_2\times W_2\times d_2}$, respectively, where $(H_1,W_1)$ and $(H_2,W_2)$ represent the spatial dimension at different layers, and $d_1$, $d_2$ represent the corresponding feature dimension. The shallow features $F_{mid}$ capture low-level spatial patterns and color information at higher spatial resolution, while $F_{deep}$ focuses on global semantic content at lower spatial resolution but with richer semantic representation. To effectively balance shallow spatial information with deep semantic knowledge while handling scale discrepancies, we use $F_{mid}$ as auxiliary information to enhance the overall feature representation. The fused hierarchical features $F_{fuse}\in\mathbb{R}^{H_2\times W_2\times d}$ are computed as follows:
\begin{equation}
    F_{fuse} = F_{deep} + \mathcal{G}_1(\mathcal{A}(F_{mid}))+\mathcal{G}_2(F_{deep}),
\end{equation}
where $\mathcal{A}(\cdot)$ is an adaptive pooling or interpolation operation that spatially aligns $F_{mid}$ to match the dimensions of $F_{deep}$ (transforming from $H_1\times W_1$ to $H_2\times W_2$). The function $\mathcal{G}_1$ encodes information from the intermediate layer feature $F_{mid}$, and $\mathcal{G}_2$ further refines the information from $F_{deep}$. Both $\mathcal{G}_1$ and $\mathcal{G}_2$ are Multi-Layer Perception (MLPs).

Subsequently, the hierarchical fused features $F_{fuse}$ are reshaped into patch tokens $F\in \mathbb{R}^{(H_2\cdot W_2)\times d}$. To aggregate object-level information and extract fine-grained class-specific features, we associate each class with a learnable embedding, forming a sequence $S\in \mathbb{R}^{M\times d}$ with $M=|\mathcal{C}^{1:t}|$ known classes at session $t$. The feature purification module consists of $L$ multi-head self-attention (MSA) blocks that jointly processes $S$ and the hierarchical feature tokens $F$, producing purified class features $O_{S}\in \mathbb{R}^{M\times d}$ and enhanced patch features $O_{F}\in \mathbb{R}^{(H_2\cdot W_2)\times d}$ (the mini-batch is omitted):
\begin{align}
    (Q,K,V)&=(W_{q},W_{k},W_{v})[F,S], \\
    O &= W_{o}\mathrm{softmax}\left ( \frac{QK^{T}}{\sqrt{d/h} }  \right )V+ b_{o},
\end{align}
where $O=[O_{F},O_{S}]$ and $h$ is the number of attention heads. Under the attention formulation, each class embedding acts as a query that selectively activates the spatial regions in $F_{fuse}$ relevant to that class and captures contextual relations with other classes, benefiting from both low-level spatial details and high-level semantic cues. This architecture naturally supports incremental tasks by appending new class embeddings, enabling efficient parallel prediction of both old and new classes with enhanced multi-scale discriminative power. 

Class features $O_S$ are then fed to the classifier to produce output logits. Following established practices~\cite{yan2021dynamically}, we employ a stability classifier to predict logits for previous classes and a plasticity classifier to generate logits for new classes, which are then concatenated to form complete logits $P^{1:t}$ for final classification. During learning new classes, we freeze the old embeddings $S^{1:t-1}\in \mathbb{R}^{|\mathcal{C}^{1:t-1}|\times d}$ and the stability classifier to preserve existing knowledge, while allowing new class embeddings $S^{t}\in \mathbb{R}^{|\mathcal{C}^{t}|\times d}$ and the plasticity classifier to adapt to incoming data. Upon transitioning to session $t+1$, the plasticity and stability classifiers consolidated to form the updated stability classifier, and a fresh plasticity classifier is instantiated for the subsequent learning.

\begin{figure}[t]
  \centering
  \includegraphics[width=0.88\linewidth]{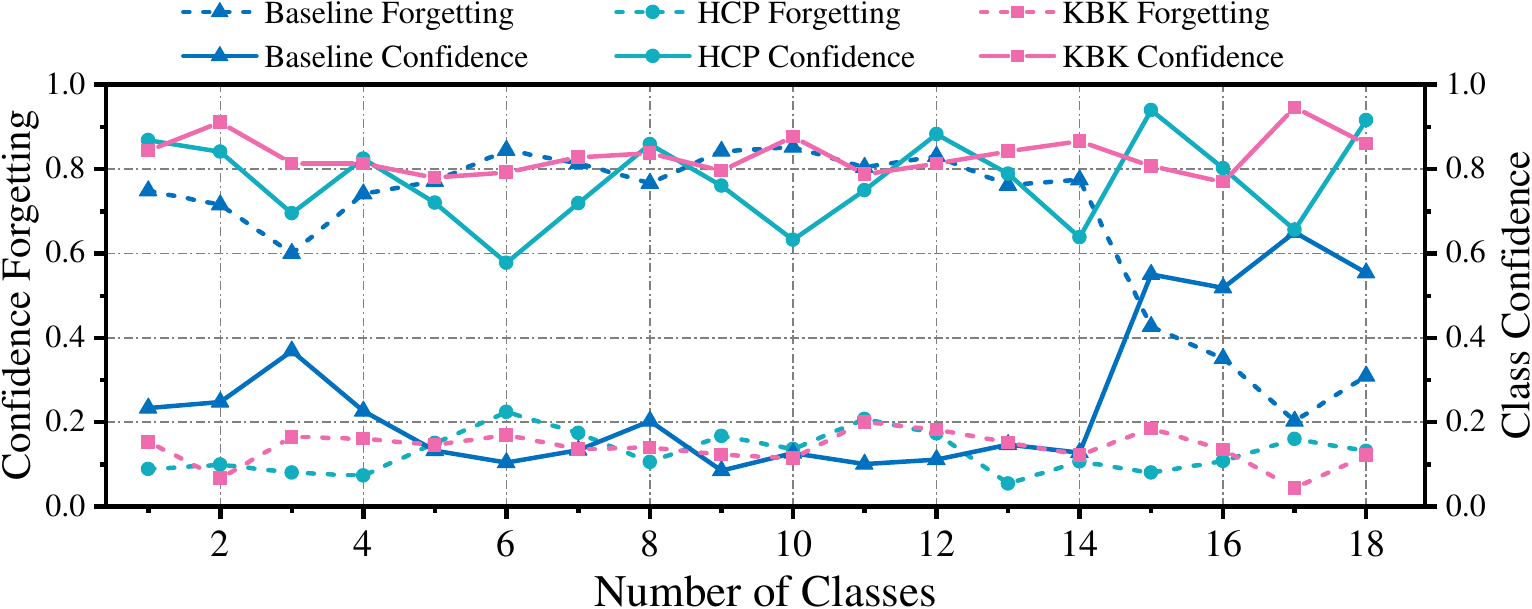}
  \vspace{-4mm}
  \caption{Heterogeneous forgetting among classes render a unified fixed threshold inadequate for recalling previously learned knowledge.}
  \label{fig:class-confidence}
  \vspace{-3mm}
\end{figure}
{\textit{2) Uncertainty-aware Recall Enhancement.}}

Specifically, leveraging the historical model's discriminative ability for past classes, we obtain the predicted class probability for old classes as $P_{i}^{t-1}=[p_{i1}^{t-1},\cdots,p_{i|1:C_{t-1}|}^{t-1}]$. At the same time, we characterize the prediction uncertainty for class $j$ using the binary entropy:
\begin{equation}
    h_{ij}^{t-1}=-p_{ij}^{t-1}\log{p_{ij}^{t-1}}-(1-p_{ij}^{t-1})\log{(1-p_{ij}^{t-1})}.
\end{equation}

Different classes, however, exhibit varying levels of learning difficulties and susceptibility to forgetting. Some categories have clear, easily distinguishable features and are thus relatively resistant to forgetting, whereas others are more subtle and consequently more prone to being forgotten. Treating all categories uniformly ignores this heterogeneity, which introduces noise to pseudo-labeling, resulting in either false positives or missed true positives. To quantify such heterogeneity, we perform a statistical analysis of confidence forgetting across classes. Taking VOC dataset under B10-C2 setting as an example, we approximate the confidence distribution of each class with a Gaussian distribution $p_{j}^{t}\sim \mathcal{N}(\mu_{j}^{t} ,\sigma^{2t}_{j} )$, where:
\begin{align}
    \mu_j^{t} &= \frac{\sum_{i=1}^{|D^{t}|} p_{ij}^{t} \cdot \mathbb{I}_{(y_{ij}^{t} = 1)}}{\sum_{i=1}^{|D^{t}|}\mathbb{I}_{(y_{ij}^{t} = 1)}},\\
    \sigma_j^{2t}&=\frac{\sum_{i=1}^{|D^{t}|} \left( p_{ij}^{t} \cdot \mathbb{I}_{(y_{ij}^{t} = 1)} - \mu_j^{t} \right)^2}{\sum_{i=1}^{|D^{t}|} \mathbb{I}_{(y_{ij}^{t} = 1)}},
\end{align}
where $\mathbb{I}(\cdot)$ denotes the indicator function, which returns 1 if the condition is true, and 0 otherwise. Following the definition of accuracy forgetting in SLCIL, we extend this notion to define class-level \textit{confidence forgetting} as the decline between the historical maximum mean confidence and the post-session mean confidence: 
\begin{align}
    CF_{j}^{t}=\max_{m\le t-1} (\mu_{j}^{m} - \mu_{j}^{t} ).
\end{align}

Empirical evidence (see Fig.~\ref{fig:class-confidence}) indicates that the baseline experiences dramatic confidence decay (up to 0.85), and that the degree of decline differs substantially across categories. This heterogeneity renders a single global threshold ineffective. To address this, we employ per-class adaptive thresholds and introduce entropy gating as a second-level filter to suppress high-uncertainty recalls. Specifically, the confidence threshold $\varepsilon_{j}$ of class $j$ is derived from the confidence statistics of previous models (e.g., mean or a robust percentile over positives), while the entropy threshold $\tau_{j}$ is estimated from the entropy distribution of reliable samples. With both thresholds taken as the mean (by default), we derive the class-specific pseudo labels to supervise training in session $t$:
\begin{equation}
    \hat{y}_{ij}^{t}=\left\{\begin{matrix}
  1,& \text{if } y_{ij}^{t}=1\text{ and } j\in \mathcal{C}^{t}, \\
  1,& \text{if }  y_{ij}^{t}=0\text{ and }j\in \mathcal{C}^{1:t-1}\text{ and }\\
    & p_{ij}^{t-1}\ge \varepsilon_{j}\text{ and }h_{ij}^{t-1} \le \tau_{j},\\
  0,& \text{otherwise} .
\end{matrix}\right.
\end{equation}

Both the confidence and entropy distribution queues are updated after each session to account for distribution drift. This uncertainty-aware class-adaptive strategy provides a clean recall signal under partial labeling, consistently improving historical precision-recall while mitigating confirmation bias.





\subsection{Probing Unknown Knowledge}
Incorporating auxiliary classes during incremental sessions helps preserve embedding capacity for upcoming categories and improve forward compatibility. Introducing real classes from external datasets is a simple option but is often infeasible in practice due to access limitations and domain shift that could disrupt the current training process. In multi-label settings, where unlabeled co-occurring classes naturally exist, we instead exploit implicit knowledge to synthesize prospective features that embody relational semantics, thereby enriching the feature representation space. By encouraging known-class embeddings to diverge from these synthesized unknown features, the model forms tighter, more discriminative clusters for known categories while implicitly reserving representational capacity for subsequent categories.

Formally, for an image $x_i$ from $\mathcal{D}^{t}$ with pseudo target $\hat{Y}_i^{t}\!=\![\hat{y}_{i1}^{t},\cdots,\hat{y}_{iM}^{t}]\!\in\!\mathbb{R}^{1\times M}$ where $M\!=\!|\mathcal{C}^{1:t}|$, we obtain purified class features $O_{S}\!\in\! \mathbb{R}^{|\mathcal{C}^{1:t}|\times d}$ via the hierarchical feature purification module. We partition these into present-class features $O_{S}^{+}$ and absent-class features $O_{S}^{-}$:
\begin{align}
    O_{S}^{+} &= \left \{ o_{j}|\hat{y}_{ij}^{t}=1,{j}\in \mathcal{C}^{1:t} \right \} \in \mathbb{R}^{M^{+}\times d},\\
    O_{S}^{-} &= \left \{ o_{j}|\hat{y}_{ij}^{t}=0,{j}\in \mathcal{C}^{1:t} \right \} \in \mathbb{R}^{M^{-}\times d},
\end{align}
with $M^{+}+M^{-}\!=\!M$. Here, the attention of present-class embeddings is supervised to focus on fine-grained target regions, whereas the absent-class embeddings distribute more freely across unexplained foreground areas, implicitly encoding signals of latent unknown categories. To exploit this implicit information, we leverage semantic relations to synthesize unknown features. Instead of random mixing, we weight absent-class features according to their semantic similarities with present classes, ensuring that synthesized features better capture the relational structure between known and potential future categories. Specifically, for each present-class embedding $o_{p}\in O_{S}^{+}$, we compute its cosine similarity with every absent-class embedding $o_{a}\in O_{S}^{-}$:
\begin{equation}
    s(o_p,o_a)=\frac{o_p\cdot o_a}{\left \| o_p \right \| \left \| o_a \right \| }.
\end{equation}

We aggregate these similarities across all present classes to produce a reliability weight for each absent class:
\begin{equation}
    w_a = \frac{1}{|O_{S}^{+}|}\sum_{o_p\in O_{S}^{+}}s(o_p,o_a).
\end{equation}

{Here, the cosine similarity is used as a contextual reliability prior rather than a predictor of the exact semantics of future classes. A larger similarity indicates that the absent-class feature is more related to the current scene, while a weakly correlated absent feature receives a smaller weight and is less likely to dominate the synthesized unknown feature.} These weights modulate a random interpolation coefficient $\lambda_a$ sampled from a Beta distribution, yielding the synthesized unknown feature $O_V$:
\begin{align}
    \lambda_{a}\sim\text{Beta}(\alpha,&\beta),\, \bar{\lambda}_{a}=\frac{\lambda_a\cdot w_a}{\sum_{j}\lambda_j\cdot w_j},\\
     O_V&=\sum_{o_a\in O_{S}^{-}}\bar{\lambda}_{a}\cdot o_a.
\end{align}

{The synthesized feature \(O_v\) is incorporated into training by extending the plasticity classifier's output to \(C^t+1\), where the extra dimension represents an explicit unknown class. Treating synthesized features as a dedicated unknown class enables the model to account for label uncertainty within the current session, sharpen discrimination among known classes, and improve preparedness for future classes.}

{It should be noted that SPU does not aim to construct exact prototypes for all future classes. Instead, the synthesized unknown feature serves as a coarse known/unknown boundary regularizer. When an absent-class feature has low correlation with the present-class context, its contribution is naturally suppressed by the similarity weighting, reducing the risk of feature-space distortion. If most absent features are weakly correlated with the current scene, the synthesized unknown feature may become less informative, and the forward-compatibility benefit of SPU may be weakened. Nevertheless, the unknown branch still helps prevent unexplained co-occurring content from being directly absorbed into known-class decision regions.}

\subsection{Category-balanced Gradient Compensation}
To address the negative-positive sample imbalance inherent to multiple scenarios, we adopt the Asymmetric Loss (ASL) as optimization objective--a focal-loss variant that applies distinct $\gamma$ for positive and negative examples. For an image $x_i$, we obtain its class probabilities $P_{i}^{t}=[p_{i1}^{t},\cdots,p_{iK}^{t}]\in\mathbb{R}^{1\times K}$, where $K=|\mathcal{C}^{1:t}|+1$ including an additional unknown class. We train the framework with ASL:
\begin{equation}
    L_{cls} =  \frac{1}{B}\frac{1}{K}\sum_{i=1}^{B} \sum_{j=1}^{K} \left\{ \begin{array}{ll}
(1 - p_{ij}^{t})^{\gamma+} \log(p_{ij}^{t}), & \text{if } \hat{y}_{ij}^{t} = 1, \\
(p_{ij}^{t})^{\gamma-} \log(1 - p_{ij}^{t}), & \text{if } \hat{y}_{ij}^{t} = 0,
\end{array} \right.
\end{equation}
where $\gamma+$ and $\gamma-$ are used to modulate the contribution of positives and negatives and set $\gamma^{+}=0$, $\gamma^{-}=4$ by default.

However, during training on incoming classes, the current dataset becomes biased toward new categories, which in turn skews the back-propagated gradients at the final layer and exacerbates catastrophic forgetting. Moreover, standard gradient optimization neglects heterogeneous forgetting spaces across old classes:  easy-to-forget classes with diverse appearances versus hard-to-forget classes with distinctive attributes. To overcome heterogeneous forgetting from the gradient perspective, we design a category-balanced gradient compensation loss that adaptively rescales gradient contributions according to each class’s forgetting pace. Specifically, the gradient update value $\mathcal{G}_{ij}^{t}$ with respect to the $j$-th neuron $\mathcal{N}_{j}^{t}$ in the last classifier layer is expressed as:
\begin{equation}
    \mathcal{G}_{ij}^{t} = \frac{\partial L_{cls}(p_{ij}^{t},\hat{y}_{ij}^{t} )}{\partial \mathcal{N}_{j}^{t} }=p_{ij}^{t}-\hat{y}_{ij}^{t}.
\end{equation}

To equalize disparate forgetting rates across easily- and hard-to-forget old classes while moderating the learning speed for newly introduced categories, we normalize gradients separately for groups of classes learned in different incremental tasks and use these normalization factors to reweight the standard classification loss $L_{cls}$. Given a mini-batch $\left \{ x_{i}^{t},\hat{y}_{i}^{t} \right \}_{i=1}^{B}$ sampled from task $t$, we compute task-wise average gradient magnitudes that have been learned in the $m$-th ($1\le m \le t$) incremental task as:
\begin{equation}
     \mathcal{G}^{m^{+}}=\frac{1}{\sum_{i=1}^{B}\sum_{j\in \mathcal{C}^{m}}\mathbb{I}_{(\hat{y}_{ij}^{t}=1)}  }\sum_{i=1}^{B}\sum_{j\in \mathcal{C}^{m}}\left | \mathcal{G}_{ij}^{t}  \right | \cdot \mathbb{I}_{(\hat{y}_{ij}^{t}=1 {)}}.
    \label{eq: gradient}
\end{equation}

Multi-label data are inherently imbalanced, where positives are scarce and negatives are abundant. Estimating the reference solely from positive samples yields high-variance, unstable normalization when positives are rare, which can spuriously inflate the weights. Incorporating negatives reduces variance, suppresses noise amplification, and is equally crucial for correcting hard negatives. Accordingly, we compute separate batch-only mean gradient magnitudes for positives and negatives per old class and choose the corresponding mean as the normalization baseline conditioned on the sample's label. This category-aware design symmetrically calibrates false negatives and false positives, improving stability and old-class retention. Thus, we rewrite Eq.(\ref{eq: gradient}) as follows:
\begin{align}
    \mathcal{G}^{m} &= \mathcal{G}^{m^+}\cdot \mathbb{I}_{(\hat{y}_{ij}^{t}=1\wedge j\in{\mathcal{C}^{m})}} + \mathcal{G}^{m^-}\cdot\mathbb{I}_{(\hat{y}_{ij}^t=0\wedge j\in{\mathcal{C}^{m})}},\\
    \mathcal{G}^{m^-}&=\frac{1}{\sum_{i=1}^{B}\sum_{j\in \mathcal{C}^{m}}\mathbb{I}_{(\hat{y}_{ij}^{t}=0)}  }\sum_{i=1}^{B}\sum_{j\in \mathcal{C}^{m}}\left | \mathcal{G}_{ij}^{t}  \right | \cdot \mathbb{I}_{(\hat{y}_{ij}^{t}=0)}.
\end{align}

Based on the task-level average gradients of old classes, we derive a category-balanced weight $\psi_{ij}^{t}$ by normalizing the per-sample gradient of class $j$ in image $i$:
\begin{equation}
    \psi _{ij}^{t}=\begin{cases}
  \frac{\left | \mathcal{G}^{t}_{ij} \right |  }{\mathcal{G}^m }, & \text{if } j\in \mathcal{C}^m \, \text{and } m< t,  \\
  1, & \text{otherwise}.
\end{cases}
\label{eq: psi_weight}
\end{equation}

The dynamically updated weight $\psi_{ij}^{t}$ is applied to reweight the classification objective, yielding the category-balanced gradient compensation loss:
\begin{equation}
    L_{gc} = \frac{1}{B}\frac{1}{K}  \sum_{i=1}^{B}\sum_{j=1}^{K}\psi_{ij}^{t}\cdot \left\{\begin{array}{ll}
(1 - p_{ij}^{t})^{\gamma+} \log(p_{ij}^{t}), &\text{if }\hat{y}_{ij}^{t} = 1, \\
(p_{ij}^{t})^{\gamma-} \log(1 - p_{ij}^{t}), &\text{if }\hat{y}_{ij}^{t} = 0.
\end{array} \right.
\label{eq: gradient_loss}
\end{equation}

Intuitively, a forgotten class typically produces large gradients $\mathcal{G}_{ij}^{t}$, which inflates $\psi_{ij}^{t}$ in Eq.(\ref{eq: psi_weight}). The larger weight increases the contribution of that class to $L_{gc}$, driving its output probability $p_{ij}^{t}$ toward the pseudo-label and thereby helping to reduce catastrophic forgetting.







\subsection{{Mechanism Discussion}}
{From the knowledge-specification perspective, partial labeling creates three coupled risks: historical classes lack explicit supervision, current classes are learned from entangled representations, and prospective classes may be prematurely absorbed into known-class regions. KBK addresses these risks through complementary mechanisms. HFP specifies historical and current knowledge at the representation level by extracting class-aware features from co-occurring objects. URE restores reliable historical supervision by filtering old-class predictions according to class-adaptive confidence and uncertainty. SPU assigns unexplained prospective information to an explicit unknown direction, preventing it from interfering with known-class representations. Finally, CGC coordinates historical retention and current learning at the optimization level by compensating heterogeneous forgetting while moderating the learning dynamics of newly introduced classes. These components form a unified progression from representation specification and supervision restoration to boundary construction and optimization balancing.}

\section{Experiments}
\subsection{Experimental Setup and Evaluation Metrics}
\textbf{Datasets.} 
We evaluate our approach on MS-COCO 2014~\cite{lin2014microsoft} and PASCAL VOC 2007~\cite{everingham2007pascal}. MS-COCO contains 82,081 training examples and 40,137 test examples across 80 common object classes ($\approx$ 2.9 labels/image). PASCAL VOC includes 5,011 images in the train-val split and 4,952 test images covering 20 classes ($\approx$ 1.6 labels/image).

\textbf{Protocols.} 
{We evaluate KBK under two protocols. Protocol A, adopted from~\cite{dong2023knowledge}, follows the standard VOC/COCO setting, where images are collected according to current-session classes and may overlap across sessions. Protocol B (following~\cite{song2024overcoming}) follows the stricter Split-VOC/Split-COCO setting, where disjoint image splits are used to avoid cross-session image overlap. Both protocols retain the key MLCIL difficulty of partial supervision, where only current classes are annotated while historical and prospective classes may appear without labels. Their different session constructions allow us to evaluate the robustness of KBK to protocol changes. Notably, since Protocol B constructs disjoint image splits before assigning class blocks, some sessions may contain few or no positive instances for certain assigned classes. This does not affect comparison fairness because all methods follow the same fixed protocol, but it makes Protocol B a stricter benchmark with irregular session composition.}

Following previous work~\cite{buzzega2020dark}, we use the unified B$i$-C$j$ convention, where $i$ denotes the number of classes learned in the base session and $j$ is the number of classes to be learned in each subsequent session. Under protocol A, experiments are conducted with MS-COCO (B40-C10, B0-C10) and Pascal VOC (B10-C2, B0-C4, B4-C2, B5-C3). For protocol B, we evaluate the model on the Split-COCO dataset with B40-C10 and B0-C20 settings and on the Split-VOC dataset with B10-C5 and B0-C5 settings.
\begin{table}[htbp]
\caption{Performance on MS-COCO. Comparison methods are grouped by their source task. A buffer size of 0 denotes no rehearsal. Best scores per setting are shown in \textbf{bold} and second-best are indicated by \underline{underline}.}
\vspace{-4mm}
\setlength{\tabcolsep}{0.1mm}
\small
\begin{center}
{
\renewcommand{\arraystretch}{0.72}
\begin{tabular}{l|c|c|cccc|cccc}
\toprule
\multirow{3}{*}{{Method}} & \multirow{3}{*}{\begin{tabular}[c]{@{}c@{}}{Source}\\ {Task}\end{tabular}} & \multirow{3}{*}{\begin{tabular}[c]{@{}c@{}}{Buffer}\\ {Size}\end{tabular}} & \multicolumn{4}{c|}{{MS-COCO B0-C10}}                           & \multicolumn{4}{c}{{MS-COCO B40-C10}}                          \\ \cline{4-11} 
                        &                                                                        &                                                                        & \multicolumn{1}{c|}{Avg. Acc} & \multicolumn{3}{c|}{Last Acc} & \multicolumn{1}{c|}{Avg. Acc} & \multicolumn{3}{c}{Last Acc} \\ \cline{4-11} 
                        &                                                                        &                                                                        & \multicolumn{1}{c|}{mAP}      & CF1      & OF1      & mAP      & \multicolumn{1}{c|}{mAP }      & CF1      & OF1     & mAP      \\ \midrule
Upper-bound             & Baseline                                                               & -                                                                      & \multicolumn{1}{c|}{-}        & 76.4     & 79.4     & 81.8    & \multicolumn{1}{c|}{-}        & 76.4     & 79.4    & 81.8    \\ \midrule \midrule
FT                      & Baseline                                                               & \multirow{8}{*}{0}                                                     & \multicolumn{1}{c|}{38.3}     & 6.1      & 13.4     & 16.9    & \multicolumn{1}{c|}{35.1}     & 6.0      & 13.6    & 17.0    \\
PODNet~\cite{douillard2020podnet}                  & SLCIL                                                                    &                                                                        & \multicolumn{1}{c|}{43.7}     & 7.2      & 14.1     & 25.6    & \multicolumn{1}{c|}{44.3}     & 6.8      & 13.9    & 24.7    \\
oEWC~\cite{schwarz2018progress}                    & SLCIL                                                                    &                                                                        & \multicolumn{1}{c|}{46.9}     & 6.7      & 13.4     & 24.3    & \multicolumn{1}{c|}{44.8}     & 11.1     & 16.5    & 27.3    \\
LWF~\cite{li2017learning}                     & SLCIL                                                                    &                                                                        & \multicolumn{1}{c|}{47.9}     & 9.0      & 15.1     & 28.9    & \multicolumn{1}{c|}{48.6}     & 9.5      & 15.8    & 29.9    \\
AGCN~\cite{du2023multi}                     & MLCIL                                                                  &                                                                        & \multicolumn{1}{c|}{72.4}     & 53.9     & 56.6     & 61.4    & \multicolumn{1}{c|}{73.9}     & 58.7     & 59.9    & 69.1    \\
KRT~\cite{dong2023knowledge}                     & MLCIL                                                                  &                                                                        & \multicolumn{1}{c|}{74.6}     & 55.6     & 56.5     & 65.9    & \multicolumn{1}{c|}{77.8}     & {64.4}     & 63.4    & 74.0    \\
CSC~\cite{du2024confidence} & MLCIL   &                                                                        & \multicolumn{1}{c|}{\underline{78.0}}     & \underline{64.9}    & \underline{66.8}     & \underline{72.8}    & \multicolumn{1}{c|}{78.2}     & \underline{65.7}     & 67.0    & 74.0 \\ 
    HCP~\cite{zhang2025specifying}                   & MLCIL                                                                  &                                                                        & \multicolumn{1}{c|}{77.9}         &  60.4        &  65.3        &  71.2& \multicolumn{1}{c|}{\underline{78.9}}         &   64.9       & \underline{68.6}        &  \underline{75.3}\\ 
\textbf{KBK} & MLCIL                                                                  &                                                                       & \multicolumn{1}{c|}{ {\textbf{80.7}}}         &   \textbf{67.3}    & \textbf{68.2}        &    {\textbf{75.8}}& \multicolumn{1}{c|}{ {\textbf{79.7}}}       &  {\textbf{69.6}} &  {\textbf{70.0}}         &   {\textbf{77.2}}\\ \midrule

TPCIL~\cite{tao2020topology} & SLCIL                                                                    & \multirow{8}{*}{5/class}                                               & \multicolumn{1}{c|}{63.8}     & 20.1     & 21.6     & 50.8    & \multicolumn{1}{c|}{63.1}     & 25.3     & 25.1    & 53.1    \\
PODNet~\cite{douillard2020podnet} & SLCIL                                                                    &                                                                        & \multicolumn{1}{c|}{65.7}     & 13.6     & 17.3     & 53.4    & \multicolumn{1}{c|}{65.4}     & 24.2     & 23.4    & 57.8    \\
DER++~\cite{buzzega2020dark} & SLCIL                                                                    &                                                                        & \multicolumn{1}{c|}{68.1}     & 33.3     & 36.7     & 54.6    & \multicolumn{1}{c|}{69.6}     & 41.9     & 43.7    & 59.0    \\
AGCN~\cite{du2023multi} & MLCIL                                                                  &                                                                        & \multicolumn{1}{c|}{72.9}     & 56.7     & 58.5     & 63.6    & \multicolumn{1}{c|}{74.5}     & 59.8     & 61.3    & 69.7    \\
KRT~\cite{dong2023knowledge} & MLCIL                                                                  &                                                                        & \multicolumn{1}{c|}{75.8}     & 60.0     & 61.0     & 68.3    & \multicolumn{1}{c|}{78.0}     & 66.0     & 65.9    & 74.3    \\
CSC~\cite{du2024confidence} & MLCIL                                                                  &                                                                        & \multicolumn{1}{c|}{79.2}     & 67.3     & 68.1     & 73.7    & \multicolumn{1}{c|}{78.4}     & 68.1     & 69.0    & 75.6    \\ 
HCP~\cite{zhang2025specifying} & MLCIL                                                                  &                                                                        & \multicolumn{1}{c|}{\underline{79.4}}         &   \underline{70.3}       & \underline{72.9}         &   \underline{74.5} & \multicolumn{1}{c|}{\underline{79.4}}         &   \underline{71.5}       &  \underline{74.1}       &  \underline{76.7}\\ 
\textbf{KBK} & MLCIL                                                                  &                                                                        & \multicolumn{1}{c|}{\textbf{{81.0}}}         &   \textbf{72.8}      & \textbf{75.7}         &   \textbf{{76.3}}& \multicolumn{1}{c|}{\textbf{{79.9}}}         &   \textbf{73.9}       &  \textbf{76.4}      &  \textbf{{77.7}}\\ \midrule

iCaRL~\cite{rebuffi2017icarl} & SLCIL                                                                    & \multirow{11}{*}{20/class}                                              & \multicolumn{1}{c|}{59.7}     & 19.3     & 22.8     & 43.8    & \multicolumn{1}{c|}{65.6}     & 22.1     & 25.5    & 55.7    \\
BiC~\cite{wu2019large}                     & SLCIL                                                                    &                                                                        & \multicolumn{1}{c|}{65.0}     & 31.0     & 38.1     & 51.1    & \multicolumn{1}{c|}{65.5}     & 38.1     & 40.7    & 55.9    \\
ER~\cite{riemer2018learning}                      & SLCIL                                                                    &                                                                        & \multicolumn{1}{c|}{60.3}     & 40.6     & 43.6     & 47.2    & \multicolumn{1}{c|}{68.9}     & 58.6     & 61.1    & 61.6    \\
TPCIL~\cite{tao2020topology}                   & SLCIL                                                                    &                                                                        & \multicolumn{1}{c|}{69.4}     & 51.7     & 52.8     & 60.6    & \multicolumn{1}{c|}{72.4}     & 60.4     & 62.6    & 66.5    \\
PODNet~\cite{douillard2020podnet}                  & SLCIL                                                                    &                                                                        & \multicolumn{1}{c|}{70.0}     & 45.2     & 48.7     & 58.8    & \multicolumn{1}{c|}{71.0}     & 46.6     & 42.1    & 64.2    \\
DER++~\cite{buzzega2020dark}                   & SLCIL                                                                    &                                                                        & \multicolumn{1}{c|}{72.7}     & 45.2     & 48.7     & 63.1    & \multicolumn{1}{c|}{73.6}     & 51.5     & 53.5    & 66.3    \\
AGCN~\cite{du2023multi}                  & MLCIL                                                                  &                                                                        & \multicolumn{1}{c|}{73.2}     & 59.5     & 60.3     & 66.0    & \multicolumn{1}{c|}{75.2}     & 64.1     & 65.2    & 71.7    \\
KRT~\cite{dong2023knowledge}                  & MLCIL                                                                  &                                                                        & \multicolumn{1}{c|}{76.5}     & 63.9     & 64.7     & 70.2    & \multicolumn{1}{c|}{78.3}     & 67.9     & 68.9    & 75.2    \\ 
CSC~\cite{du2024confidence} & MLCIL & & \multicolumn{1}{c|}{79.6} & 67.8  & 68.6 & \underline{74.8} & \multicolumn{1}{c|}{78.7} & 68.2 & 69.4 & 76.7 \\
{HCP}~\cite{zhang2025specifying}                     & MLCIL                                                                  &                                                                        & \multicolumn{1}{c|}{\underline{79.6}}         &   \underline{70.4}       &     \underline{73.0}     &  74.6& \multicolumn{1}{c|}{\underline{79.6} }        &         \underline{71.9} &         \underline{74.5} & \underline{77.2}\\
\textbf{KBK} & MLCIL                                                                  &                                                                        & \multicolumn{1}{c|}{\textbf{{81.4}}}         &   \textbf{73.5}       & \textbf{76.3}        &   \textbf{{77.2}}& \multicolumn{1}{c|}{\textbf{{80.4}}}         &   \textbf{74.4}       &  \textbf{76.7}       &  \textbf{{78.3}}\\
\midrule

PRS~\cite{kim2020imbalanced}                     & MLOIL                                                                  & \multirow{7}{*}{1000}                                                  & \multicolumn{1}{c|}{48.8}     & 8.5      & 14.7     & 27.9    & \multicolumn{1}{c|}{50.8}     & 9.3      & 15.1    & 33.2    \\
OCDM~\cite{liang2022optimizing}                    & MLOIL                                                                  &                                                                        & \multicolumn{1}{c|}{49.5}     & 8.6      & 14.9     & 28.5    & \multicolumn{1}{c|}{51.3}     & 9.5      & 15.5    & 34.0    \\
AGCN~\cite{du2023multi}                   & MLCIL                                                                  &                                                                        & \multicolumn{1}{c|}{73.0}     & 59.4     & 65.9     & 59.0    & \multicolumn{1}{c|}{75.0}     & 63.1     & 64.8    & 71.1    \\
KRT~\cite{dong2023knowledge}                   & MLCIL                                                                  &                                                                        & \multicolumn{1}{c|}{75.7}     & 61.6     & 63.6     & 69.3    & \multicolumn{1}{c|}{78.3}     & 67.5     & 68.5    & 75.1    \\
CSC~\cite{du2024confidence}                   & MLCIL                                                                  &                                                                        & \multicolumn{1}{c|}{79.3}     & 67.5     & 68.5     & 73.9    & \multicolumn{1}{c|}{78.5}     & 67.8     & 69.7    & 76.0    \\
HCP~\cite{zhang2025specifying}                    & MLCIL                                                                  &                                                                        & \multicolumn{1}{c|}{\underline{79.5}}         &   \underline{70.2}      &   \underline{72.8}       &         \underline{74.4} & \multicolumn{1}{c|}{\underline{79.5}}         &   \underline{71.8}       &  \underline{74.4}       & \underline{76.7}        \\ 
\textbf{KBK} & MLCIL                                                                  &                                                                        & \multicolumn{1}{c|}{\textbf{{81.6}}}         &   \textbf{73.7}       & \textbf{76.5}         &   \textbf{{77.3}}& \multicolumn{1}{c|}{\textbf{{80.3}}}         &   \textbf{74.3}       &  \textbf{76.8}      &  \textbf{{78.2}}\\
\bottomrule
\end{tabular}
}
\end{center}
\label{ms-coco-results}
\vspace{-3mm}
\end{table}
\begin{table}[t]
\caption{Comparison of the results on PASCAL VOC dataset. Data in \textbf{Bold} represents the best results.}
\label{voc-results}
\setlength{\tabcolsep}{0.6mm}
\small
\begin{center}
{
\renewcommand{\arraystretch}{0.72}
\begin{tabular}{l|c|cc|cc|cc|cc}
\toprule
\multirow{2}{*}{Method} & \multirow{2}{*}{\begin{tabular}[c]{@{}c@{}}Buffer\\ Size\end{tabular}} & \multicolumn{2}{c|}{{VOC B0-C4}} & \multicolumn{2}{c|}{{VOC B10-C2}}& \multicolumn{2}{c|}{{VOC B5-C3}}& \multicolumn{2}{c}{{VOC B4-C2}} \\ \cline{3-10} 
                        &                                                                        & Avg.        & Last      & Avg.       & Last   & Avg.        & Last & Avg.        & Last      \\ \midrule
Upper bound             &  -                                               & -              & 93.6          & -              & 93.6&-               & 93.6& -              & 93.6          \\ \midrule
FT             &   \multirow{5}{*}{0}                                                                    & 82.1           & 62.9          & 72.9           & 43.0  &   74.4 & 49.4 & 60.4 & 37.0     \\ 
KRT~\cite{dong2023knowledge} & &89.1 & 80.2 & 84.3 & 71.5  &84.1 & 72.1 & 67.5 & 43.5 \\
CSC~\cite{du2024confidence} & & 90.4 & 85.1 & 89.0 & \underline{83.8} &88.0 & 82.1 & \underline{83.3} & \textbf{74.1}\\
HCP~\cite{zhang2025specifying} && \underline{92.9} & \underline{87.9}&\underline{90.1}& {81.9}& \underline{89.6} &\underline{82.9} & 82.9 & 70.1\\
\textbf{KBK} & & \textbf{94.1} & \textbf{90.9} & \textbf{90.9} & \textbf{85.0} & \textbf{90.8} & \textbf{84.8} &\textbf{85.5} &\underline{72.2}\\\midrule
iCarL~\cite{rebuffi2017icarl}                   & \multirow{10}{*}{2/class}                                               & 87.2           & 72.4          & 79.0           & 66.7  &   - & -&-&-       \\
BiC~\cite{wu2019large}                     &                                                                        & 86.8           & 72.2          & 81.7           & 69.7 &    - & -&-&-         \\
ER~\cite{riemer2018learning}                      &                                                                        & 86.1           & 71.5          & 81.5           & 68.6  &     79.0 & 65.3 & 73.7 & 57.9   \\
TPCIL~\cite{tao2020topology}                   &                                                                        & 87.6           & 77.3          & 80.7           & 70.8  &    - & -&-&-       \\
PODNet~\cite{douillard2020podnet}                  &                                                                        & 88.1           & 76.6          & 81.2           & 71.4 &    81.4 & 70.3 & 75.7 & 62.5    \\
DER++~\cite{buzzega2020dark}                   &                                                                        & 87.9           & 76.1          & 82.3           & 70.6 &    78.0 & 68.1 & 77.0 & 61.6     \\
KRT~\cite{dong2023knowledge}                   &                                                                        & 90.7           & 83.4          & 87.7           & 80.5    & 89.4 & 82.5 & 82.0 & 72.6    \\
CSC~\cite{du2024confidence}                   &                                                                        & 92.4           & 87.9          & 91.6           & \underline{87.8} & \underline{91.9} & \underline{87.5} & \textbf{90.4}&\textbf{86.6}         \\
HCP~\cite{zhang2025specifying}                      &                                                                        &  {\underline{93.5}}          &    {\underline{89.2}}           &       \underline{92.1}         &  {86.3}  &   91.3 & 83.9 & 85.5 & 77.7        \\ 
\textbf{KBK}                      &                                                                        &  \textbf{94.7}          &    \textbf{91.3}           &       \textbf{92.8}        &  \textbf{88.8}   & \textbf{92.5} & \textbf{87.8} & \underline{90.1} & \underline{81.8}          \\ 
\bottomrule
\end{tabular}
}
\end{center}
\vspace{-4mm}
\end{table}

\textbf{Evaluation Metrics.}
Following KRT~\cite{dong2023knowledge}, we use mean average precision (mAP) as the primary metric, computed over all classes learned up to each session. We summarize mAP with two indicators: (1) average mAP, the arithmetic mean of mAP values across sessions, which reflects overall continual performance; and (2) last mAP, the mAP observed at the final session. To give a complete picture of multi-label performance, we also report the per-class F1 score (CF1) and the overall F1 score (OF1).

\subsection{Implementation Details}
{For fair comparison, we follow the general training settings of prior MLCIL works, especially KRT. Each experiment is run three times under the same protocol, and the average performance is reported}. The model is trained for 20 epochs with a batch size of 64 using Adam, weight decay \(1\times10^{-4}\), and the OneCycleLR schedule. The learning rate is \(4\times10^{-5}\) for the base session, and \(1\times10^{-4}\) / \(4\times10^{-5}\) for incremental sessions on MS-COCO / PASCAL VOC, respectively. All experiments are conducted on NVIDIA RTX 3090 GPUs. In HFP, we set the number of attention heads to 4, use 3 MSA blocks for VOC and 1 for MS-COCO, and set the embedding dimension to \(d=2048\). At session \(t\), the model is initialized from the \((t-1)\)-th checkpoint; old class embeddings are fixed, while new class embeddings are randomly initialized. We merge the previous stability and plasticity classifiers into a frozen stability classifier for \(|\mathcal{C}^{1:t-1}|\) historical classes, and append a plasticity classifier with \(|\mathcal{C}^{t}|+1\) outputs for current classes and the unknown class. {For SPU, the interpolation coefficient is sampled from \(\mathrm{Beta}(0.2,0.2)\). For URE, the class-adaptive confidence and entropy thresholds are estimated from historical prediction distributions and updated after each session.}

\subsection{Main Results}
\textbf{Results on MS-COCO.}
Tab.~\ref{ms-coco-results} exposes severe forgetting in simple fine-tuning (FT) and SLCIL methods: FT's Last Acc collapses to 16.9\% and PODNet to 25.6\%, whereas KBK achieves 75.8\% on B0-C10 with zero buffer. Multi-label online incremental learning methods likewise achieve weak results here, yet our method substantially outperforms PRS and OCDM. Compared with the latest MLCIL methods, our method consistently maintains a leading position, achieving up to 2.3\% higher Avg. Acc than CSC (buffer size=1000) and clear metric-wise improvements relative to AGCN. Stable gains under the B40-C10 setting further highlight the robustness of KBK. Notably, even in a no-replay setting, it surpasses competing methods that rely on memory buffers in both average and final-session accuracy. {For a fair interpretation, we note that KBK, KRT, KRT-R, and HCP are compared under the same TResNetM backbone, which forms the strictly backbone-matched comparison. In contrast, L2P and DualPrompt are based on pre-trained ViT backbones with larger parameter scales, and are therefore included only as cross-architecture references rather than strictly same-backbone baselines.}
Tab.~\ref{tab:prompt-compare} illustrates the comparison between KBK and prompt-based methods, which shows that KBK achieves higher performance across all scenarios with the same parameters as KRT.

\begin{table}[t]
\caption{{Comparison with prompt-based methods on MS-COCO. KBK, KRT, KRT-R, and HCP use the same TResNetM backbone, while L2P and DualPrompt are ViT-based prompt methods and are included as cross-architecture references.}}
\label{tab:prompt-compare}
\vspace{-2mm}
\setlength{\tabcolsep}{1.5mm}
\small
\begin{center}
{
\renewcommand{\arraystretch}{0.72}
\begin{tabular}{l|c|cc|cc}
\toprule
\multirow{2}{*}{{Method}} & \multirow{2}{*}{Param.} & \multicolumn{2}{c|}{{COCO B0-C10}} & \multicolumn{2}{c}{{COCO B40-C10}} \\ \cline{3-6} 
                        &                                                                        & Avg.       & Last      & Avg.       & Last \\ \midrule
Upper bound  & \multirow{5}{*}{86.0M} &-&83.16& -                                                    & 83.16                                                \\
L2P\cite{wang2022learning}                                  &                        & 73.28 & 67.72 & 73.07 & 70.42                                                \\
L2P-R\cite{wang2022learning}                                  &                        &     73.87 & 68.22 & 73.64 & 71.68                                                                                             \\ 
Dual-prompt~\cite{wang2022dualprompt}  & & 74.67 & 69.39 & 74.45 & 71.95\\
Dual-prompt-R~\cite{wang2022dualprompt} & &74.87 & 70.20 & 74.54 & 72.60\\ \midrule
Upper bound  & \multirow{5}{*}{29.4M} & -                                                    & 81.80  &-&81.80                                              \\
KRT~\cite{dong2023knowledge}  & &74.64 & 65.94 & 77.83 & 74.02\\
KRT-R~\cite{dong2023knowledge}                                  &                        & 76.53 & 71.24 & 78.34 & 75.18                                                \\
HCP~\cite{zhang2025specifying}                                 &        & \underline{79.61} & \underline{74.60}              & \underline{79.64}                                                & \underline{77.18}                                                \\ 
\textbf{KBK}  & & \textbf{81.36} & \textbf{77.19} & \textbf{80.35} & \textbf{78.33} \\ \bottomrule
\end{tabular}
}
\end{center}
\vspace{-4mm}
\end{table}

\textbf{Results on PASCAL VOC.}
Tab.~\ref{voc-results} reports the performance comparison on VOC under four different protocols. In the buffer-free setting, KBK sets a new state of the art across all settings. Specifically, under B0-C4, it reaches an Avg. Acc of 94.1\%, outperforming both previous best method CSC (90.4\%) and HCP (92.9\%). In the more challenging long-sequence B10-C2, KBK attains an Avg. Acc of 90.9\%, demonstrating strong stability in long-term incremental learning. Particularly under VOC B4-C2, it significantly exceeds other methods with a 2.6\% improvement over HCP, highlighting its scalability under high-frequency class increments. When a small exemplar buffer (2 samples/class) is available, KBK continues to maintain a leading position, achieving 92.8\% and 92.5\% in Avg. Acc under B10-C2 and B5-C3, respectively, with Last Acc also clearly exceeding competing methods (88.8\% and 87.8\%), confirming its ability to effectively leverage limited memory for enhanced knowledge retention and predictive consistency in multi-label class-incremental learning. 

\begin{table}[t]
\caption{Performance on Split-VOC and Split-COCO datasets. The input size is 448 × 448, and all results are obtained from APPLE~\cite{song2024overcoming}.}
  \vspace{-1mm}
\label{tab:split_results}
\setlength{\tabcolsep}{0.5mm}
\scriptsize
\begin{center}
{
\renewcommand{\arraystretch}{0.72}
\resizebox{\textwidth}{!}{
\begin{tabular}{l|c|cc|cc|c|cc|cc}
\toprule
\multirow{2}{*}{{Method}} & \multirow{2}{*}{\begin{tabular}[c]{@{}c@{}}{Buffer}\\ {Size}\end{tabular}} & \multicolumn{2}{c|}{{Split-VOC B10-C5}} & \multicolumn{2}{c|}{{Split-VOC B0-C5}}& \multirow{2}{*}{\begin{tabular}[c]{@{}c@{}}{Buffer}\\ {Size}\end{tabular}}& \multicolumn{2}{c|}{{Split-COCO B40-C10}}& \multicolumn{2}{c}{{Split-COCO B0-C20}} \\ \cline{3-6} \cline{8-11} 
                        &                                                                        & Avg.        & Last       & Avg.        & Last  & &Avg.        & Last & Avg.        & Last      \\ \midrule
Upper bound             &  -                                               & -              & 94.2          & -              & 94.2&-& -              & 86.4& -              & 86.4          \\ \midrule
FT                      &        \multirow{3}{*}{0}                                                                 &  74.8& 60.1 & 74.1 &59.7 & \multirow{3}{*}{0}&35.8 &11.1 & 51.9 &23.6   \\ 
{HCP}~\cite{zhang2025specifying} && \underline{92.5} & \underline{89.2} & \underline{92.2} & \underline{86.9} && \underline{83.2} & \underline{80.4} & \underline{85.0} & \underline{81.9}\\
\textbf{KBK} & & \textbf{95.6} & \textbf{92.5} & \textbf{95.7} & \textbf{92.0} & &\textbf{84.5}&\textbf{82.5}&\textbf{87.0}&\textbf{84.3}\\\midrule
iCarL~\cite{rebuffi2017icarl}                   & \multirow{9}{*}{5/class}                                               &  90.8& 87.1 & 88.3 &84.8& \multirow{9}{*}{20/class} & 76.7 &65.6 & 76.5 &64.5     \\
ER~\cite{riemer2018learning}                      &   &      86.0 &73.9 & 82.7& 68.3 &     &72.3 &64.0 & 63.6 &50.1  \\
TPCIL~\cite{tao2020topology}                   &   &      90.2 &84.2 & 87.9& 79.4 &   & 69.4 &71.2 & 73.5& 68.9  \\
PODNet~\cite{douillard2020podnet}                  &                                                                        &   90.3 & 86.3 &  87.8&  84.1 & &  77.1 & 66.1 & 74.8&  61.0   \\
PASS~\cite{zhu2021prototype} & & 86.0 &76.9 & 75.6 & 51.8& &73.8 &59.4 & 72.2 &49.9 \\
DER++~\cite{buzzega2020dark}                   &                                                                        &     90.2 & 83.9 &  88.9 & 84.8  && 66.7 & 55.8 & 73.8 & 67.3   \\
APPLE~\cite{song2024overcoming}                   &                                                                        &  91.7 & 89.4 &  89.5 & 85.6 & & 82.1 & 74.6& 83.5 & 76.6   \\
HCP~\cite{zhang2025specifying}                      &                                                                        &       \underline{93.4} & \underline{90.6} & \underline{93.5} & \underline{89.7} & & \underline{84.0} & \underline{82.0} & \underline{85.2} & \underline{82.3}   \\ 
\textbf{KBK}                      &                                                                        & \textbf{96.1} & \textbf{93.9} & \textbf{96.2} & \textbf{93.5} && \textbf{85.4}& \textbf{83.7} & \textbf{87.2} & \textbf{84.8}  \\ 
\bottomrule
\end{tabular}}
}
\end{center}
\label{split-coco-voc}
\vspace{-4mm}
\end{table}

\begin{table}[t]
\caption{Ablation study w.r.t. Hierarchical Feature Purification (HFP), Uncertainty-aware Recall Enhancement (URE), Semantic-guided Probing Unknown (SPU), and Category-balanced Gradient Compensation (CGC) on Pascal VOC B10-C2 and B4-C2 settings.}
\vspace{-2mm}
\label{tab:ablation_main}
\setlength{\tabcolsep}{2.5mm}
\small
\begin{center}
{
\renewcommand{\arraystretch}{0.72}
\begin{tabular}{ccccc|cc|cc}
\toprule
\multirow{2}{*}{Method} & \multirow{2}{*}{FP} & \multirow{2}{*}{RE} & \multirow{2}{*}{PU} & \multirow{2}{*}{} & \multicolumn{2}{c|}{VOC B10-C2} & \multicolumn{2}{c}{VOC B4-C2} \\ \cmidrule{6-9} 
                        &&                        &                         &                     & Avg.       & Last       & Avg.       & Last      \\ \midrule
                        Baseline & & & & &72.87 &  46.76 & 60.20 & 22.01 \\
                        & \checkmark & & & & 82.09 & 66.29 & 75.88 & 52.95 \\
                        & \checkmark & \checkmark & & & 86.00 & 73.45 & 80.73 & 64.44 \\
                        & \checkmark & & \checkmark & &86.47 &71.64 & 80.30 & 62.30 \\
                       HCP & \checkmark & \checkmark & \checkmark & & \textbf{90.05} & \textbf{81.85} & \textbf{82.94} & \textbf{70.08}  \\ \midrule \midrule 
& \multirow{1}{*}{HFP} & \multirow{1}{*}{URE} & \multirow{1}{*}{SPU} & \multirow{1}{*}{CGC} & Avg. & Last & Avg. &Last \\ \midrule
                    &\checkmark&                     &     &                & 85.21 &  70.12  &78.67 & 60.47                   \\
                    &\checkmark&   \checkmark                  &  &                   & 87.01 &    75.36    & 81.30  & {\textbf{72.63}}               \\
                    &\checkmark & \checkmark & &\checkmark & 89.28 & 80.37 & 84.51 & 70.71 \\
                    &\checkmark&      \checkmark               &  \checkmark  &                 & 90.63 &  83.37 & 85.03 & 68.78 \\
                    
                    KBK &\checkmark&      \checkmark               &  \checkmark  &  \checkmark              & {{\textbf{90.89}}}  & {{\textbf{84.95}}} & {{\textbf{85.53}}}  & {72.24} \\
                    \bottomrule
\end{tabular}
}
\end{center}
\vspace{-5mm}
\end{table}

\textbf{Results on Split-COCO and Split-VOC.}
Tab.~\ref{split-coco-voc} summarizes the experimental results on the challenging Split-COCO and Split-VOC benchmarks. We observe that our method, KBK, delivers substantial gains over rehearsal-based baselines under these split protocols and advances the state of the art in MLCIL. Specifically, KBK improves Avg. Acc over APPLE by 4.4\% on B10-C5 and 6.7\% on B0-C5. Remarkably, even with no replay buffer (buffer = 0), KBK matches or exceeds many replay-based methods and substantially outperforms simple fine-tuning. These results indicate that KBK is robust to the stricter, non-overlapping session design used in the split benchmarks and can preserve and transfer prior knowledge without relying on exemplars.

{The consistent improvements across VOC, COCO, Split-VOC, and Split-COCO also indicate that KBK is not tailored to a single dataset or session construction. These benchmarks differ in data scale, label density, class composition, and co-occurrence patterns, providing evidence for the robustness of KBK.}

\begin{table}[t]
\caption{Ablation of recall strategy. Rows one and two correspond to a single global threshold, row three shows top-K filtering, RE refers to Recall Enhancement, and URE to uncertainty-aware recall enhancement.}
\vspace{-2mm}
\label{tab:pseudo-ablate}
\setlength{\tabcolsep}{3mm}
\small
\begin{center}
{
\renewcommand{\arraystretch}{0.72}
\begin{tabular}{c|cc|cc}
\toprule
\multirow{2}{*}{Recall Strategy} & \multicolumn{2}{c|}{VOC B10-C2} & \multicolumn{2}{c}{VOC B4-C2} \\ \cmidrule{2-5} 
                        &                  Avg.       & Last       & Avg.       & Last      \\ \midrule
                        $\varepsilon\!=\!0.8$ & 83.09 &  68.94 & 74.45 & 57.84 \\
                        $\varepsilon\!=\!0.9$ & 85.08 & 70.82 & 75.63 & 55.63 \\
                        Top-2 & 84.24 & 68.94 & 75.14 & 56.48 \\
                        RE & 90.05 & 81.85 & 83.02 & 65.51 \\
                       URE & \textbf{90.89} & \textbf{84.95} & \textbf{85.53} & \textbf{72.24}  \\ 
                    \bottomrule
\end{tabular}
}
\end{center}
\vspace{-5mm}
\end{table}

\begin{table}[t]
\centering
\small
\setlength{\tabcolsep}{3mm}
\setlength{\aboverulesep}{1pt}
\setlength{\belowrulesep}{1pt}
\setlength{\extrarowheight}{0pt}
\caption{Effect of freezing old class embeddings on VOC.}
\label{tab:frozen_embedding}
\begin{tabular}{l|cc|cc}
\toprule
\multirow{2}{*}{Frozen} 
& \multicolumn{2}{c|}{VOC B10-C2} 
& \multicolumn{2}{c}{VOC B0-C4} \\
\cmidrule(lr){2-3}\cmidrule(lr){4-5}
& Avg. Acc & Last Acc & Avg. Acc & Last Acc \\
\midrule
No  & 89.8 & 83.4 & 93.2 & 88.8 \\
\textbf{Yes} & \textbf{90.9} & \textbf{85.0} & \textbf{94.1} & \textbf{90.9} \\
\bottomrule
\end{tabular}
\vspace{-3mm}
\end{table}

\begin{table}[t]
\centering
\caption{Sensitivity analysis of the Beta distribution parameters in SPU.}
\small
\label{tab:beta_sensitivity}
\begin{tabular}{c|cc|cc}
\toprule
\multirow{2}{*}{$\mathrm{Beta}(\alpha,\beta)$}
& \multicolumn{2}{c|}{VOC B10-C2}
& \multicolumn{2}{c}{VOC B0-C4} \\
\cmidrule{2-5}
& Avg. Acc & Last Acc & Avg. Acc & Last Acc \\
\midrule
$\mathrm{Beta}(0.1,0.1)$ & 90.6 & 84.7 & 93.9 & 90.5 \\
$\mathrm{Beta}(0.2,0.2)$ & \textbf{90.9} & 85.0 & \textbf{94.1} & \textbf{90.9} \\
$\mathrm{Beta}(0.5,0.5)$ & 90.7 & \textbf{85.1} & 93.8 & 90.6 \\
$\mathrm{Beta}(1.0,1.0)$ & 90.3 & 84.3 & 93.6 & 89.9 \\
\bottomrule
\end{tabular}
\vspace{-3mm}
\end{table}

\begin{table}[t]
\centering
\caption{Computational efficiency comparison on VOC B10-C2.}
\label{tab:efficiency}
\small
\renewcommand{\arraystretch}{0.82}
\begin{tabular}{l|ccccc}
\toprule
Method & Params (M) & FLOPs (G) & Training Time & Memory (GB/GPU) \\
\midrule
KRT & 29.40 & 5.73 & 28min & 7.46 \\
HCP & 29.44 & 8.34 & 19min & 9.80 \\
KBK & 29.48 & 9.25 & 25min & 9.86 \\
\bottomrule
\end{tabular}
\end{table}

\begin{figure}[t]
  \centering
\includegraphics[width=0.9\linewidth]{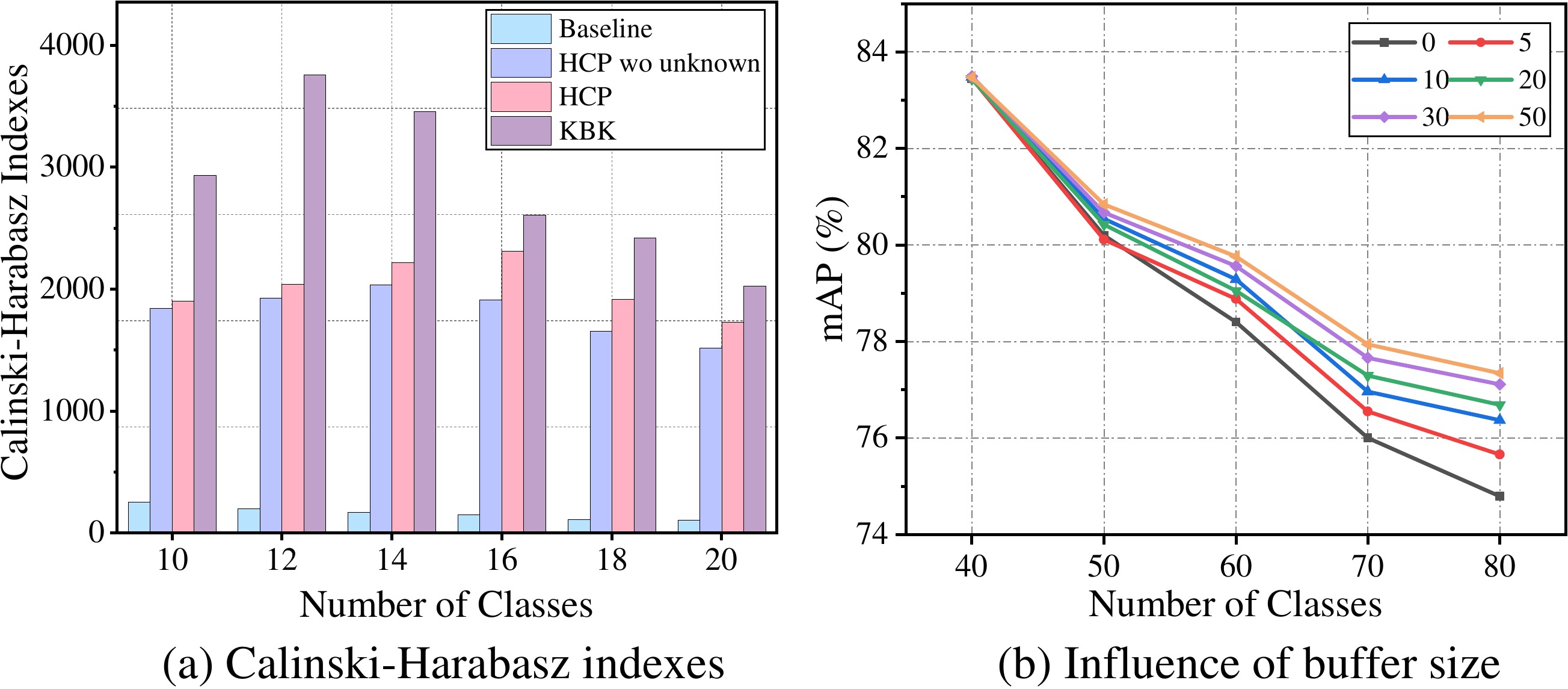}
  \vspace{-4mm}
  \caption{C-H Indexes of class features and the influence of buffer size.}
  \label{fig:CH_buffer}
  \vspace{-4mm}
\end{figure}
\begin{figure}[t]
  \centering
  \includegraphics[width=\linewidth]{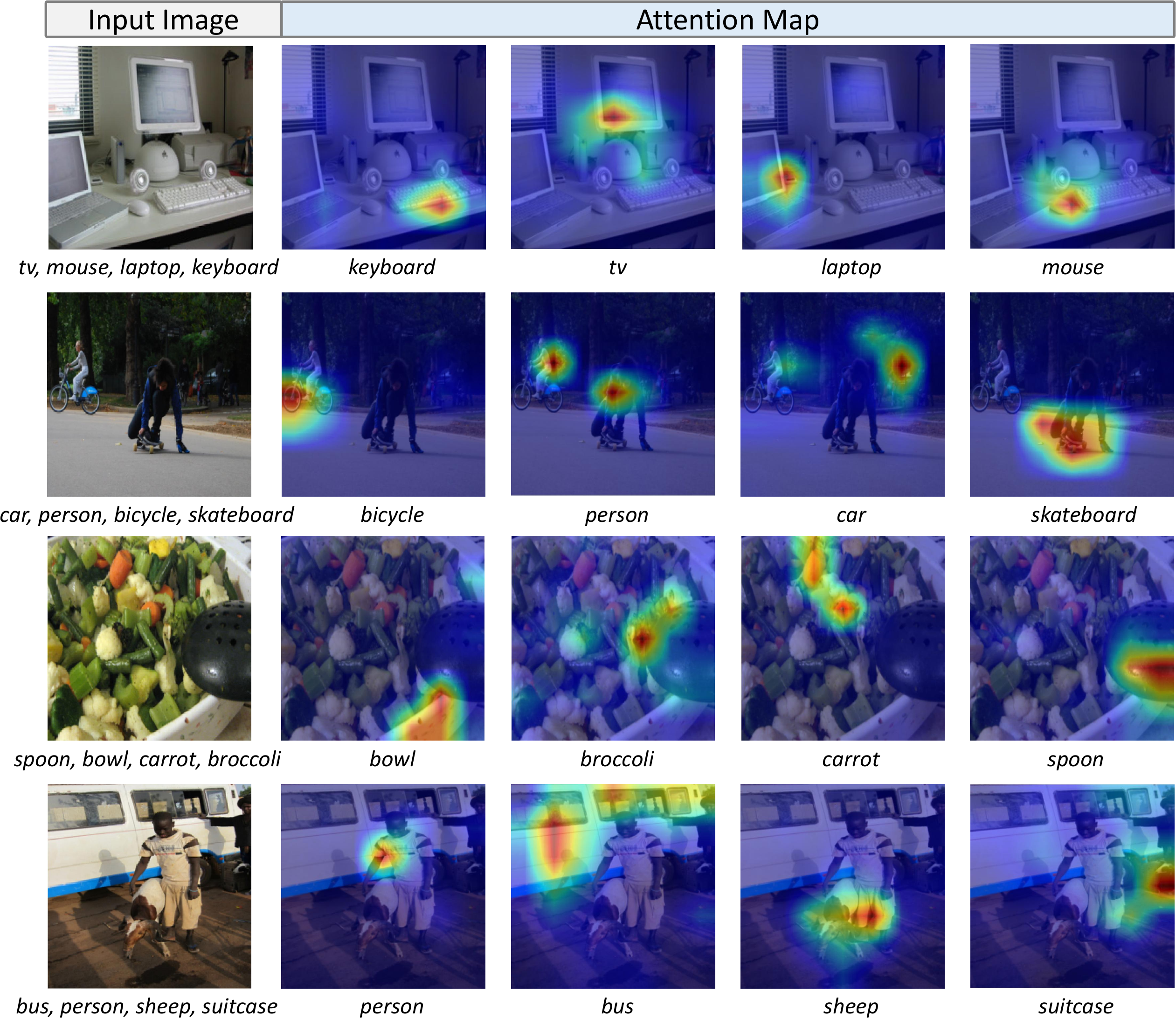}
  \vspace{-8mm}
  \caption{Visualization of attention maps for each ground-truth label in the Hierarchical Feature Purification (HFP) structure.}
  \label{fig:attention-map-voc}
  \vspace{-4mm}
\end{figure}

\begin{figure}[t]
  \centering
  \includegraphics[width=0.8\linewidth]{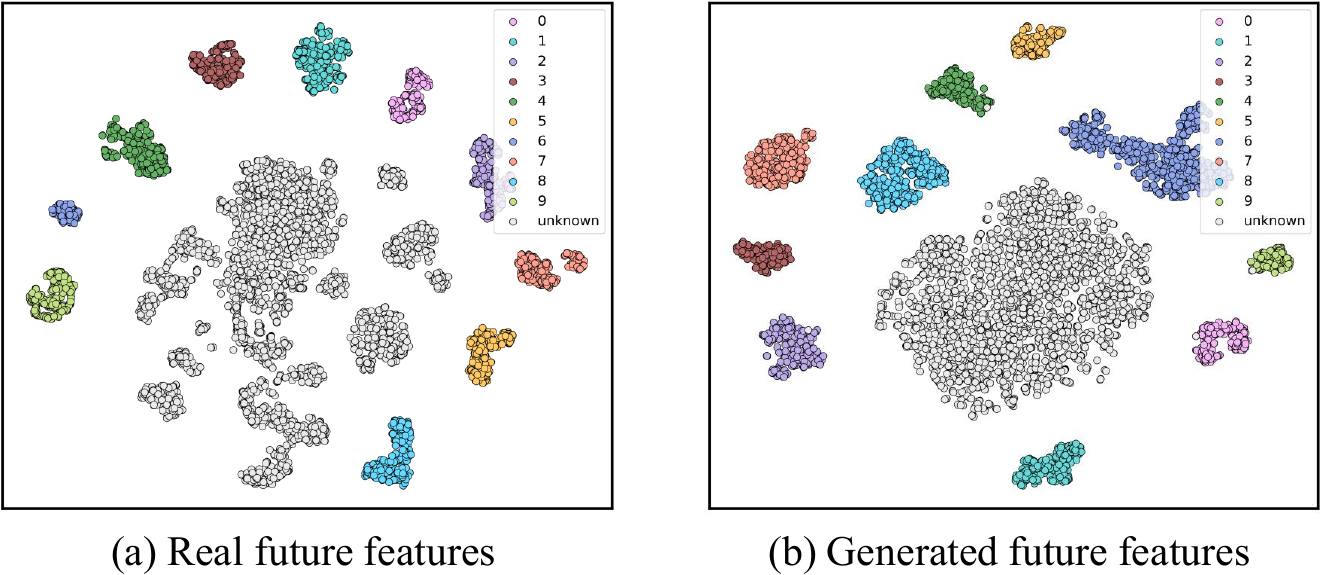}
  \vspace{-3mm}
  \caption{Visualization of real future features and generated unknown features.}
  \vspace{-3mm}
  \label{fig:virtual_index}
\end{figure}

\subsection{Ablation Study}
\textbf{Effectiveness of Component.}
{As shown in Tab.~\ref{tab:ablation_main}, fine-tuning suffers from severe forgetting, whereas FP improves Avg. Acc by 9.22 and 15.68 points on B10-C2 and B4-C2, respectively. HFP further improves Avg. Acc by 3.12 and 2.79 points by strengthening the representation-level specification of historical and current knowledge. Based on HFP, URE improves Last Acc by 5.24 and 12.16 points through more reliable historical supervision. Compared with KBK without SPU, introducing SPU improves Avg./Last Acc by 1.61/4.58 points on B10-C2 and 1.02/1.53 points on B4-C2, confirming the benefit of separating prospective information from known-class regions. Finally, CGC further improves Last Acc by 1.58 and 3.46 points by balancing historical retention and current-class optimization. Together, HFP, URE, SPU, and CGC progressively specify knowledge at the representation, supervision, feature-boundary, and optimization levels, demonstrating that KBK forms a unified framework rather than a collection of independent components.}

\begin{figure}[t]
  \centering
  \includegraphics[width=0.85\linewidth]{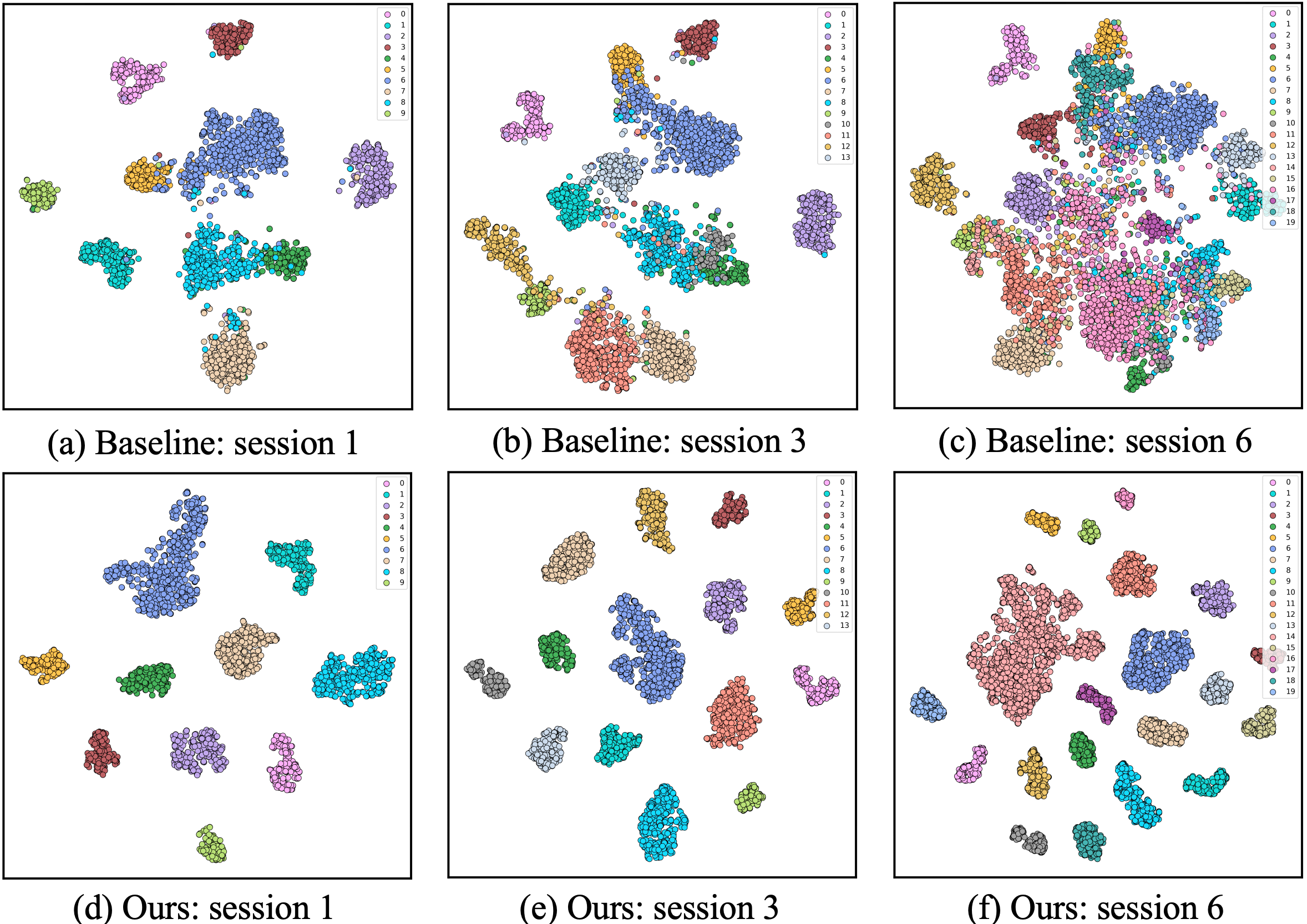}
  \vspace{-2mm}
  \caption{t-SNE visualization of feature distributions under PASCAL VOC B10-C2 setting. There are 10, 14 and 20 classes in session 1, 3 and 6, respectively.}
  \label{fig:tsne_compare}
  \vspace{-3mm}
\end{figure}

{\textbf{Effect of Freezing Old Class Embeddings.}
We further investigate the effect of freezing old class embeddings during incremental learning. As shown in Tab.~\ref{tab:frozen_embedding}, freezing old embeddings consistently improves the performance on both VOC B10-C2 and VOC B0-C4. Specifically, on VOC B10-C2, freezing improves Avg. Acc from 89.8 to 90.9 and Last Acc from 83.4 to 85.0. On VOC B0-C4, it improves Avg. Acc from 93.2 to 94.1 and Last Acc from 88.8 to 90.9. The improvement is more evident on Last Acc, indicating that freezing old embeddings helps preserve stable semantic anchors for historical classes and alleviates long-term forgetting. Meanwhile, the feature pathway, new class embeddings, and classifier remain trainable, so the model can still adapt to new classes without relying on the drift of old embeddings. These results support the stability-plasticity benefit of freezing old class embeddings in KBK.}

{\textbf{Sensitivity to Beta Distribution Parameters.}
We further analyze the sensitivity of SPU to the Beta distribution parameters used for unknown-feature synthesis. As shown in Tab.~\ref{tab:beta_sensitivity}, KBK maintains stable performance under different Beta settings. The default setting $\mathrm{Beta}(0.2,0.2)$ achieves the best or comparable results. Compared with smoother interpolation distributions such as $\mathrm{Beta}(0.5,0.5)$, $\mathrm{Beta}(0.2,0.2)$ encourages more diverse interpolation coefficients and provides stronger perturbation for unknown-feature synthesis. The overall performance variation remains small, indicating that SPU does not rely on a narrowly tuned Beta configuration.
}

{\textbf{Computational Efficiency.}
We further analyze the computational cost of KBK. As shown in Tab.~\ref{tab:efficiency}, KRT, HCP, and KBK share the same basic TResNetM backbone. HCP and KBK additionally introduce class embeddings, which add negligible additional parameter cost compared with the backbone scale. We also report FLOPs, training time, and peak GPU memory under the same hardware setting. The results show that KBK introduces only moderate computational overhead while maintaining lightweight inference. This is because the feature purification module is shared across classes and sessions, and SPU/CGC are mainly training-time operations without additional inference branches.
}

\textbf{Analysis of Buffer Size.} 
Fig.~\ref{fig:CH_buffer}(b) illustrates the results of KBK on MS-COCO B40-C10 under varying buffer sizes. Increasing the buffer from 0 to 50 yields a modest improvement in average mAP, indicating that KBK performs effectively even under strict storage constraints while still benefiting from the availability of replay buffers.

\textbf{Analysis of Hierarchical Feature Purification.}
To illustrate how KBK encodes class-specific information, we visualize the attention map between each ground-truth class embedding and the fused image tokens in the HFP module. Fig.~\ref{fig:attention-map-voc} presents the input image in column one and per-class attention overlays in the remaining columns. We observe that each class embedding can capture corresponding specific local features and locate the specified object approximately, even when similar classes co-occur in the same image. All validate that the HFP structure can extract fine-grained class-aware knowledge for learning multiple classes while avoiding class confusion.

\textbf{Ablation of Recall Enhancement.} We evaluate several pseudo-labeling schemes for recalling past knowledge: a single global threshold $\varepsilon$, top-$K$ selection, confidence-based recall enhancement (RE), and our uncertainty-aware recall enhancement (URE). Tab.~\ref{tab:pseudo-ablate} reveals that model performance depends strongly on the choice of $\varepsilon$. By contrast, URE, which discards predictions with low confidence or high uncertainty and simultaneously accounts for class-wise heterogeneous forgetting, yields more reliable supervisory signals and improved results.

\textbf{Analysis of Generated Unknown Features.}
Fig.~\ref{fig:virtual_index} compares real future features with synthetically generated unknown features (shown in gray), which both promote compact feature representations. It can be observed that the distribution of our synthesized features closely aligns with that of real future features, which confirms the effectiveness of probing unknown mechanism in providing valuable foresight for incremental learning. 
{It should be noted that SPU does not aim to predict the exact semantics of future classes in advance. Instead, it prevents unlabeled prospective-class information from being prematurely absorbed into known-class decision regions, thereby maintaining a more flexible feature space for later class expansion.}
For quantitative evaluation, Fig.~\ref{fig:CH_buffer} (a) reports the Calinski-Harabasz Index of feature representations, where higher values indicate better inter-class separation and intra-class compactness. Our method achieves significantly higher index values at each session compared to the baseline. {Moreover, removing the probing-unknown mechanism leads to a clear decline in later sessions, showing that unknown-feature synthesis is important for preserving a well-structured representation space. These results support the forward-compatibility role of SPU: by assigning uncertain prospective information to an explicit unknown direction, KBK reduces its interference with known classes and prepares for future expansion.}

\textbf{Analysis of Feature Aliasing.} 
Fig.~\ref{fig:tsne_compare} visualizes feature trajectories produced during incremental learning on VOC. In session 1, our method successfully learns pure knowledge representation without feature aliasing, enhancing the model's forward compatibility by reserving sufficient embedding space to accommodate future classes. In contrast, as training proceeds, the baseline embeddings progressively collapse and further succumb to catastrophic forgetting in session 6, manifested by blurred boundaries between previously learned classes and inability to maintain distinct representations for new and old classes. Our method consistently maintains class separation across all sessions, with each class's features forming compact and non-overlapping clusters. These results verify that our approach effectively preserves knowledge from prior sessions, adapts to current incremental tasks, and anticipates future class expansion. 

\section{Limitations}
{
Although KBK achieves consistent improvements on standard MLCIL benchmarks, several limitations remain. First, our experiments are mainly conducted on PASCAL VOC and MS-COCO, which enable fair comparison with prior MLCIL methods but cannot fully represent more complex real-world continual streams, such as long-tailed distributions, open-world category evolution, or domain shifts. Second, SPU adopts a single aggregated unknown class to regularize the known/unknown boundary, which is lightweight but cannot model fine-grained taxonomy among unseen categories. When absent-class features are weakly correlated with the current scene or dominated by background responses, the synthesized unknown feature may become less informative. Third, URE reduces noisy historical recall through confidence and uncertainty constraints, but over-confident incorrect predictions may still pass the recall gates under severe ambiguity or distribution shift. Extending KBK to more complex tasks such as incremental object detection and semantic segmentation is an interesting direction.
}

\section{Conclusion}
This paper proposes KBK, an effective framework for multi-label class-incremental learning that explicitly specifies which knowledge is known and what is unknown at each session, thereby accommodating historical, current and prospective information. For the known component, we introduce hierarchical feature purification to disentangle fine-grained class-aware features from fused semantic and visual information, which reduces feature aliasing both within and across sessions. To strengthen historical supervision under partial labeling, we study the heterogeneous forgetting among categories and further develop uncertainty-aware recall enhancement that suppresses low-confidence, high-uncertainty predictions. To probe the unknown, we exploit semantic relations to generate informative unknown features as a prospective class to enhance inter-class discrimination and reserve representational room for future additions. Furthermore, a category-balanced gradient compensation loss adaptively rescales gradient contributions to balance differing forgetting velocities among classes. Experiments and ablation studies confirm the effectiveness and robustness of KBK framework.

In future work, we will explore extending the knowledge-specification principle to more complex visual tasks such as incremental object detection and semantic segmentation. {In these tasks, known/unknown ambiguity may arise at the region or pixel level, where unlabeled objects or regions can interfere with both old and new categories. Adapting uncertainty-aware recall, prospective unknown modeling, and category-balanced optimization to localization-aware architectures remains an interesting future direction.}



\section*{Acknowledgments}
This work is supported by the National Natural Science Foundation of China (Grant NO 62406318, 62376266, 62076195, 62376070, 62406167 and U24B6012).



\bibliographystyle{elsarticle-num} 
\bibliography{hcp}

@inproceedings{yan2021dynamically,
  title={{DER}: Dynamically expandable representation for class incremental learning},
  author={Yan, Shipeng and Xie, Jiangwei and He, Xuming},
  booktitle={Proceedings of the IEEE/CVF Conference on Computer Vision and Pattern Recognition},
  pages={3014--3023},
  year={2021}
}

@inproceedings{douillard2022dytox,
  title={Dytox: Transformers for continual learning with dynamic token expansion},
  author={Douillard, Arthur and Ram{\'e}, Alexandre and Couairon, Guillaume and Cord, Matthieu},
  booktitle={Proceedings of the IEEE/CVF Conference on Computer Vision and Pattern Recognition},
  pages={9285--9295},
  year={2022}
}

@inproceedings{rebuffi2017icarl,
  title={{iCARL}: Incremental classifier and representation learning},
  author={Rebuffi, Sylvestre-Alvise and Kolesnikov, Alexander and Sperl, Georg and Lampert, Christoph H},
  booktitle={Proceedings of the IEEE/CVF conference on Computer Vision and Pattern Recognition},
  pages={2001--2010},
  year={2017}
}

@inproceedings{douillard2020podnet,
  title={{PODNet}: Pooled outputs distillation for small-tasks incremental learning},
  author={Douillard, Arthur and Cord, Matthieu and Ollion, Charles and Robert, Thomas and Valle, Eduardo},
  booktitle={Proceedings of the European Conference on Computer Vision},
  pages={86--102},
  year={2020}
}

@inproceedings{wang2022learning,
	title={Learning to prompt for continual learning},
	author={Wang, Zifeng and Zhang, Zizhao and Lee, Chen-Yu and Zhang, Han and Sun, Ruoxi and Ren, Xiaoqi and Su, Guolong and Perot, Vincent and Dy, Jennifer and Pfister, Tomas},
	booktitle={Proceedings of the IEEE/CVF Conference on Computer Vision and Pattern Recognition},
	pages={139--149},
	year={2022}
}

@inproceedings{wang2022dualprompt,
	title={{DualPrompt}: Complementary prompting for rehearsal-free continual learning},
	author={Wang, Zifeng and Zhang, Zizhao and Ebrahimi, Sayna and Sun, Ruoxi and Zhang, Han and Lee, Chen-Yu and Ren, Xiaoqi and Su, Guolong and Perot, Vincent and Dy, Jennifer and others},
	booktitle={Proceedings of the European Conference on Computer Vision},
	pages={631--648},
	year={2022},
}

@article{liang2022optimizing,
  title={Optimizing class distribution in memory for multi-label online continual learning},
  author={Liang, Yan-Shuo and Li, Wu-Jun},
  journal={arXiv preprint arXiv:2209.11469},
  year={2022}
}

@inproceedings{kim2020imbalanced,
  title={Imbalanced continual learning with partitioning reservoir sampling},
  author={Kim, Chris Dongjoo and Jeong, Jinseo and Kim, Gunhee},
  booktitle={Proceedings of the European Conference on Computer Vision},
  pages={411--428},
  year={2020}
}

@article{du2023multi,
  title={Multi-label continual learning using augmented graph convolutional network},
  author={Du, Kaile and Lyu, Fan and Li, Linyan and Hu, Fuyuan and Feng, Wei and Xu, Fenglei and Xi, Xuefeng and Cheng, Hanjing},
  journal={IEEE Transactions on Multimedia},
  year={2023}
}

@inproceedings{dong2023knowledge,
  title={Knowledge restore and transfer for multi-label class-incremental learning},
  author={Dong, Songlin and Luo, Haoyu and He, Yuhang and Wei, Xing and Cheng, Jie and Gong, Yihong},
  booktitle={Proceedings of the IEEE/CVF International Conference on Computer Vision},
  pages={18711--18720},
  year={2023}
}

@article{kirkpatrick2017overcoming,
  title={Overcoming catastrophic forgetting in neural networks},
  author={Kirkpatrick, James and Pascanu, Razvan and Rabinowitz, Neil and Veness, Joel and Desjardins, Guillaume and Rusu, Andrei A and Milan, Kieran and Quan, John and Ramalho, Tiago and Grabska-Barwinska, Agnieszka and others},
  journal={National Academy of Sciences},
  volume={114},
  number={13},
  pages={3521--3526},
  year={2017}
}

@inproceedings{schwarz2018progress,
  title={{Progress \& Compress}: A scalable framework for continual learning},
  author={Schwarz, Jonathan and Czarnecki, Wojciech and Luketina, Jelena and Grabska-Barwinska, Agnieszka and Teh, Yee Whye and Pascanu, Razvan and Hadsell, Raia},
  booktitle={Proceedings of the International Conference on Machine Learning},
  pages={4528--4537},
  year={2018}
}

@inproceedings{aljundi2018memory,
  title={Memory aware synapses: Learning what (not) to forget},
  author={Aljundi, Rahaf and Babiloni, Francesca and Elhoseiny, Mohamed and Rohrbach, Marcus and Tuytelaars, Tinne},
  booktitle={Proceedings of the European Conference on Computer Vision},
  pages={139--154},
  year={2018}
}

@article{li2017learning,
  title={Learning without forgetting},
  author={Li, Zhizhong and Hoiem, Derek},
  journal={IEEE Transactions on Pattern Analysis and Machine Intelligence},
  volume={40},
  number={12},
  pages={2935--2947},
  year={2017}
}

@article{riemer2018learning,
  title={Learning to learn without forgetting by maximizing transfer and minimizing interference},
  author={Riemer, Matthew and Cases, Ignacio and Ajemian, Robert and Liu, Miao and Rish, Irina and Tu, Yuhai and Tesauro, Gerald},
  journal={arXiv preprint arXiv:1810.11910},
  year={2018}
}

@inproceedings{wu2019large,
  title={Large scale incremental learning},
  author={Wu, Yue and Chen, Yinpeng and Wang, Lijuan and Ye, Yuancheng and Liu, Zicheng and Guo, Yandong and Fu, Yun},
  booktitle={Proceedings of the IEEE/CVF Conference on Computer Vision and Pattern Recognition},
  pages={374--382},
  year={2019}
}

@inproceedings{buzzega2020dark,
  title={Dark experience for general continual learning: a strong, simple baseline},
  author={Buzzega, Pietro and Boschini, Matteo and Porrello, Angelo and Abati, Davide and Calderara, Simone},
  booktitle={Proceedings of the Advances in Neural Information Processing Systems},
  volume={33},
  pages={15920--15930},
  year={2020}
}

@inproceedings{lin2014microsoft,
  title={Microsoft {COCO}: Common objects in context},
  author={Lin, Tsung-Yi and Maire, Michael and Belongie, Serge and Hays, James and Perona, Pietro and Ramanan, Deva and Doll{\'a}r, Piotr and Zitnick, C Lawrence},
  booktitle={Proceedings of the European Conference on Computer Vision},
  pages={740--755},
  year={2014}
}

@article{everingham2007pascal,
  title={The pascal visual object classes challenge,(voc2007) results},
  author={Everingham, Mark},
  journal={http://pascallin. ecs. soton. ac. uk/challenges/VOC/voc2007/index. html.},
  year={2007}
}

@inproceedings{tao2020topology,
  title={Topology-preserving class-incremental learning},
  author={Tao, Xiaoyu and Chang, Xinyuan and Hong, Xiaopeng and Wei, Xing and Gong, Yihong},
  booktitle={Proceedings of the European Conference on Computer Vision},
  pages={254--270},
  year={2020}
}

@InProceedings{yang2022multi,
  title={Multi-view correlation distillation for incremental object detection},
  author={Yang, Dongbao and Zhou, Yu and Zhang, Aoting and Sun, Xurui and Wu, Dayan and Wang, Weiping and Ye, Qixiang},
  booktitle={Pattern Recognition},
  volume={131},
  pages={108863},
  year={2022}
}

@inproceedings{shmelkov2017incremental,
  title={Incremental learning of object detectors without catastrophic forgetting},
  author={Shmelkov, Konstantin and Schmid, Cordelia and Alahari, Karteek},
  booktitle={Proceedings of the IEEE/CVF International Conference on Computer Vision},
  pages={3400--3409},
  year={2017}
}

@inproceedings{yang2023one,
  title={One-shot replay: Boosting incremental object detection via retrospecting one object},
  author={Yang, Dongbao and Zhou, Yu and Hong, Xiaopeng and Zhang, Aoting and Wang, Weiping},
  booktitle={Proceedings of the AAAI Conference on Artificial Intelligence},
  volume={37},
  number={3},
  pages={3127--3135},
  year={2023}
}

@inproceedings{li2023patchct,
  title={{PatchCT}: Aligning patch set and label set with conditional transport for multi-label image classification},
  author={Li, Miaoge and Wang, Dongsheng and Liu, Xinyang and Zeng, Zequn and Lu, Ruiying and Chen, Bo and Zhou, Mingyuan},
  booktitle={Proceedings of the IEEE/CVF International Conference on Computer Vision},
  pages={15348--15358},
  year={2023}
}

@inproceedings{zhang2025specifying,
  title={Specifying What You Know or Not for Multi-Label Class-Incremental Learning},
  author={Zhang, Aoting and Yang, Dongbao and Liu, Chang and Hong, Xiaopeng and Zhou, Yu},
  booktitle={Proceedings of the AAAI Conference on Artificial Intelligence},
  volume={39},
  number={21},
  pages={22345--22353},
  year={2025}
}

@inproceedings{zhang2025dca,
  title={{DCA}: Dividing and Conquering Amnesia in Incremental Object Detection},
  author={Zhang, Aoting and Yang, Dongbao and Liu, Chang and Hong, Xiaopeng and Shang, Miao and Zhou, Yu},
  booktitle={Proceedings of the AAAI Conference on Artificial Intelligence},
  volume={39},
  number={9},
  pages={9851--9859},
  year={2025}
}

@inproceedings{song2024overcoming,
  title={Overcoming catastrophic forgetting for multi-label class-incremental learning},
  author={Song, Xiang and Shu, Kuang and Dong, Songlin and Cheng, Jie and Wei, Xing and Gong, Yihong},
  booktitle={Proceedings of the IEEE/CVF Winter Conference on Applications of Computer Vision},
  pages={2389--2398},
  year={2024}
}

@inproceedings{du2024confidence,
  title={Confidence self-calibration for multi-label class-incremental learning},
  author={Du, Kaile and Zhou, Yifan and Lyu, Fan and Li, Yuyang and Lu, Chen and Liu, Guangcan},
  booktitle={Proceedings of the European Conference on Computer Vision},
  pages={234--252},
  year={2024},
}

@inproceedings{zhu2021prototype,
  title={Prototype augmentation and self-supervision for incremental learning},
  author={Zhu, Fei and Zhang, Xu-Yao and Wang, Chuang and Yin, Fei and Liu, Cheng-Lin},
  booktitle={Proceedings of the IEEE/CVF Conference on Computer Vision and Pattern Recognition},
  pages={5871--5880},
  year={2021}
}

@inProceedings{Douillard_2021_CVPR,
    author    = {Douillard, Arthur and Chen, Yifu and Dapogny, Arnaud and Cord, Matthieu},
    title     = {{PLOP}: Learning Without Forgetting for Continual Semantic Segmentation},
    booktitle = {Proceedings of the IEEE/CVF Conference on Computer Vision and Pattern Recognition},
    year      = {2021},
    pages     = {4040-4050}
}

@inProceedings{Michieli_2021_CVPR,
    author    = {Michieli, Umberto and Zanuttigh, Pietro},
    title     = {Continual Semantic Segmentation via Repulsion-Attraction of Sparse and Disentangled Latent Representations},
    booktitle = {Proceedings of the IEEE/CVF Conference on Computer Vision and Pattern Recognition},
    year      = {2021},
    pages     = {1114-1124}
}

@inProceedings{oh2022alife,
  title={Alife: Adaptive logit regularizer and feature replay for incremental semantic segmentation},
  author={Oh, Youngmin and Baek, Donghyeon and Ham, Bumsub},
  booktitle={Proceedings of the Advances in Neural Information Processing Systems},
  volume={35},
  pages={14516--14528},
  year={2022}
}

@article{yang2022rd,
  title={{RD-IOD}: Two-level residual-distillation-based triple-network for incremental object detection},
  author={Yang, Dongbao and Zhou, Yu and Shi, Wei and Wu, Dayan and Wang, Weiping},
  journal={ACM Transactions on Multimedia Computing, Communications, and Applications},
  volume={18},
  number={1},
  pages={1--23},
  year={2022},
}

@article{liu2025class,
  title={Class incremental learning with self-supervised pre-training and prototype learning},
  author={Liu, Wenzhuo and Wu, Xin-Jian and Zhu, Fei and Yu, Ming-Ming and Wang, Chuang and Liu, Cheng-Lin},
  journal={Pattern Recognition},
  volume={157},
  pages={110943},
  year={2025},
}

@article{dong2023class,
  title={Class-incremental object detection},
  author={Dong, Na and Zhang, Yongqiang and Ding, Mingli and Bai, Yancheng},
  journal={Pattern Recognition},
  volume={139},
  pages={109488},
  year={2023},
}

@article{sun2023exemplar,
  title={Exemplar-free class incremental learning via discriminative and comparable parallel one-class classifiers},
  author={Sun, Wenju and Li, Qingyong and Zhang, Jing and Wang, Danyu and Wang, Wen and Geng, YangLi-ao},
  journal={Pattern Recognition},
  volume={140},
  pages={109561},
  year={2023},
}

@inproceedings{chen2024modeling,
  title={Modeling Label Correlations with Latent Context for Multi-label Recognition},
  author={Chen, Zhaomin and Cui, Quan and Deng, Ruoxi and Hu, Jie and Zhang, Guodao},
  booktitle={European Conference on Computer Vision},
  pages={218--234},
  year={2024},
}

@article{zhu2024semantic,
  title={Semantic-guided representation enhancement for multi-label image classification},
  author={Zhu, Xuelin and Li, Jianshu and Cao, Jiuxin and Tang, Dongqi and Liu, Jian and Liu, Bo},
  journal={IEEE Transactions on Circuits and Systems for Video Technology},
  volume={34},
  number={10},
  pages={10036--10049},
  year={2024},
}

@article{chen2024pursuit,
  title={In pursuit of causal label correlations for multi-label image recognition},
  author={Chen, Zhao-Min and Jin, Xin and Ge, Yisu and Chan, Sixian},
  journal={Advances in Neural Information Processing Systems},
  volume={37},
  pages={51634--51654},
  year={2024}
}




\end{document}